\documentclass{article}

\pdfoutput=1 

\usepackage[colorlinks,citecolor=blue,urlcolor=blue]{hyperref}
\usepackage{graphicx}%
\usepackage{multirow}%
\usepackage{amsmath,amssymb,amsfonts}%
\usepackage{amsthm}%
\usepackage{mathrsfs}%
\usepackage[title]{appendix}%
\usepackage{xcolor}%
\usepackage{textcomp}%
\usepackage{booktabs}%
\usepackage{algorithm}%
\usepackage{algorithmicx}%
\usepackage{algpseudocode}%
\usepackage{listings}%
\usepackage{url}

\usepackage[authoryear]{natbib}

\usepackage{bm}
\usepackage{authblk}

\usepackage[margin=1.2in]{geometry}

\newcommand{\rcode}[1]{\normalfont\texttt{#1}}

\newcommand{\myheading}[1]{\par\medskip\noindent\textit{#1}\par\medskip\noindent\ignorespaces}

\newbox{\myorcidaffilbox}
\sbox{\myorcidaffilbox}{\large\includegraphics[height=1.7ex]{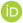}}
\newcommand{\orcidaffil}[1]{%
  \href{https://orcid.org/#1}{\usebox{\myorcidaffilbox}}}

\DeclareMathOperator*{\argmax}{arg\,max}

\title{Multi forests: Variable importance for multi-class outcomes}
\author[1,2,*]{Roman Hornung  \orcidaffil{0000-0002-6036-1495}}
\author[3]{Alexander Hapfelmeier \orcidaffil{0000-0001-6765-6352}}
\affil[1]{Institute for Medical Information Processing, Biometry and Epidemiology, LMU Munich, Munich, Germany}
\affil[2]{Munich Center for Machine Learning (MCML), Munich, Germany}
\affil[3]{Institute of AI and Informatics in Medicine, TUM School of Medicine and Health, Technical University of Munich, Munich, Germany}
\affil[*]{Corresponding author: Roman Hornung, hornung@ibe.med.uni-muenchen.de}

\begin{document}

\maketitle

\begin{abstract}
In prediction tasks with multi-class outcomes, identifying covariates specifically associated with one or more outcome classes can be important. Conventional variable importance measures (VIMs) from random forests (RFs), like permutation and Gini importance, focus on overall predictive performance or node purity, without differentiating between the classes. Therefore, they can be expected to fail to distinguish class-associated covariates from covariates that only distinguish between groups of classes. We introduce a VIM called multi-class VIM, tailored for identifying exclusively class-associated covariates, via a novel RF variant called multi forests (MuFs). The trees in MuFs use both multi-way and binary splitting. The multi-way splits generate child nodes for each class, using a split criterion that evaluates how well these nodes represent their respective classes. This setup forms the basis of the multi-class VIM, which measures the discriminatory ability of the splits performed in the respective covariates with regard to this split criterion. Alongside the multi-class VIM, we introduce a second VIM, the discriminatory VIM. This measure, based on the binary splits, assesses the strength of the general influence of the covariates, irrespective of their class-associatedness. Simulation studies demonstrate that the multi-class VIM specifically ranks class-associated covariates highly, unlike conventional VIMs which also rank other types of covariates highly. Analyses of 121 datasets reveal that MuFs often have slightly lower predictive performance compared to conventional RFs. This is, however, not a limiting factor given the algorithm's primary purpose of calculating the multi-class VIM.
\end{abstract}

\section{Introduction}

Random forests (RFs) \citep{Breiman:2001} are among the most widely used classical machine learning methods for covariate-based prediction of outcomes. They exhibit strong predictive performance, can capture complex relationships between covariates and the outcome, and consider only the order of the covariate values, making them well suited in the presence of various kinds of covariate distributions. Additionally, as demonstrated in \citet{Probst:2019}, they typically provide good prediction results even without the optimization of tuning parameters.

A further likely reason for the popularity of RFs is the variable importance measures (VIMs) associated with them, which enable ranking of the covariates according to the strength of their influence on the outcome. For simplicity, we use the term \lq\lq influence'' here, but actually refer to the observed correlation between the covariates and the outcome. It is important to note that VIMs typically do not employ causal inference techniques (see e.g., \citet{Pearl:2009}), which is why they do not allow to determine whether covariates have a causal impact on the outcome. Throughout the rest of the paper, we will often use \lq\lq influence'' or \lq\lq influential'' in this simplified context, indicating the observed correlation, which may not necessarily imply a causal relationship.

Notably, the permutation VIM, already introduced in \citet{Breiman:2001}, was highlighted in \citet{Molnar:2020} as a milestone development in interpretable machine learning. Alongside the permutation VIM, the Gini importance has long been a popular VIM, primarily due to its lower computational demand. However, concerns about the use of the Gini importance in the context of CART trees were expressed as early as in \citet{Breiman:1984}. Later, \citet{Strobl:2007} demonstrated that this measure tends to overestimate the importance of covariates with many potential split points. In recent years, several bias-corrected versions of the Gini importance have been developed. For example, \citet{Nembrini:2018} adjust the Gini importance using permuted versions of the covariates, whereas \citet{Li:2019}, \citet{Loecher:2020}, and \citet{Zhou:2021} use the out-of-bag (OOB) data from the trees to achieve unbiased Gini importance estimates.

In two-class classification problems, influential continuous covariates have value ranges with distinct regions predominantly containing observations of one class over the other. This facilitates interpretations like: smaller covariate values likely indicate class $A$ and larger values likely indicate class $B$. In multi-class classification scenarios, which are the focus of this paper, the situation is more complex. Here, a covariate can have a strong influence without value regions associated with specific classes. Such a covariate might only distinguish well between two or three class groups, but may not be uniquely associated with single classes.

It is likely, and is supported by the simulation studies in this paper, that conventional VIMs assign high importance values to covariates irrespective of their specific association with classes of the outcome. This is problematic in applications where identifying covariates associated with distinct classes is of interest. Conventional VIMs assess overall predictive performance or the purity of child nodes produced by the splits, without differentiating between the classes of the outcome. Furthermore, these VIMs are based on trees that use binary splitting. In each split, the observations in the respective node are divided into two child nodes based on a split point in the values of a covariate. Here, it is likely that a different covariate is used in the next split. Therefore, this type of splitting does generally not facilitate finding subdivisions of the values of the covariates where several intervals mainly contain observations of specific classes. Multi-way splitting can overcome this limitation by dividing each node into more than two child nodes, each representing adjacent intervals in the covariates used for splitting.

This paper proposes a VIM specifically designed for multi-class outcomes, the \emph{multi-class VIM}, alongside a variant of RFs, \emph{multi forests} (MuFs), which use multi-way splitting tailored for multi-class outcomes. The multi-class VIM aims to prioritize covariates that exhibit regions substantially containing observations of specific classes, with each region being unique to a different class. Throughout the remainder of this paper, the term \lq\lq class-associated covariates'' will frequently be used to refer to these types of covariates. Covariates that merely differentiate among groups of classes without clear class-specific associations should be ranked lower than class-associated covariates in applications in which it is of interest to specifically identify covariates that are associated with one or more classes of the outcome. Note that the purpose of developing the MuF algorithm was mainly to realize the multi-class VIM. MuFs were not expected to have better predictive performance than conventional RFs.

MuFs perform $C$-way splits in (quasi-)continuous covariates or categorical covariates with at least $C$ categories (see Section \ref{sec:mufs} for details), assigning a different one of the $C$ classes of the outcome to each of the $C$ child nodes. The split criterion rewards regions within the splitting covariate's range substantially containing observations from single classes. In contrast, the criterion is negatively affected by regions with overlapping class-specific distributions of the covariate without a particular class being overrepresented. Similar to in the case of the Gini importance, this criterion is also used for calculating the multi-class VIM, exclusively using OOB data to avoid overfitting biases akin to those found with the Gini importance. Additionally, a second VIM based on the Gini importance, the \emph{discriminatory VIM}, is introduced that evaluates the discriminatory ability of binary splits. This measure can be used to also select covariates that can only distinguish well between groups of classes without being specifically associated with individual classes, as is the case also with conventional VIMs. At each node in the trees of MuFs, a random decision determines whether the split will be multi-way or binary. This integration of binary splits also addresses the issue of rapid data division by multi-way splits, discussed in Chapter~9.2 of \citet{Hastie:2009}, which can compromise the predictive performance of the forest.

The remainder of this paper is organized as follows: Section~\ref{sec:relwork} provides a brief overview of related work, specifically on decision tree approaches that use multi-way splitting and covariate selection methods for multi-class outcomes. Section~\ref{sec:mufs} details the algorithm for training and predicting with MuFs. The approaches for computing the multi-class VIM and the discriminatory VIM are covered in Section~\ref{sec:vims}. Section~\ref{sec:simstudy} presents the design and results of a simulation study, which focuses on comparing the multi-class VIM with conventional VIMs in terms of their ability to prioritize class-associated covariates. Section~\ref{sec:rdanalysis} discusses the results of a benchmark study based on 121 datasets, comparing the predictive performance of MuFs against that of conventional RFs, which serve as the reference method. The discussion in Section~\ref{sec:discussion} summarizes the proposed methodology, the results obtained, and addresses further related points. Finally, Section~\ref{sec:conclusions} succinctly summarizes the main conclusions of the paper.

\section{Related work}
\label{sec:relwork}

To the best of our knowledge, no decision tree approaches in the existing literature specifically use multi-way splitting to model multi-class outcomes. Consequently, there are also no such approaches tailored to estimate VIMs aimed at identifying class-associated covariates.

There are also relatively few decision tree approaches that employ multi-way splitting at all. Notable among these are \citet{Fayyad:1993} and \citet{Yen:2007}, which generate multi-way splits by repeated binary splitting in the same covariate. \citet{Berzal:2004} form the intervals for the multi-way splits through hierarchical clustering, selecting the subdivision within the hierarchy that minimizes a node impurity criterion. In contrast, the algorithm introduced by \citet{Elomaa:1999} facilitates the identification of multi-way splits that are optimal among all possible multi-way splits, provided that the used evaluation metric meets a specific criterion for well-behavedness. Generally speaking, all algorithms that produce multi-way splits in decision trees can be characterized as supervised discretization techniques. These techniques convert continuous covariates into discrete ones while considering their association with the outcome. For a comprehensive review of discretization techniques, refer to \citet{Garcia:2013}.

Furthermore, to the best of our knowledge, there are no other VIMs specifically designed for multi-class outcomes in the literature. Nevertheless, there are several covariate selection methods for multi-class outcomes. Examples include \citet{Forman:2004}, \citet{Chen:2006}, \citet{Zini:2015}, and \citet{Song:2017}. These methods typically focus either on the differences between each individual class and the other classes, or on the differences among all classes.

\citet{Forman:2004} and \citet{Chen:2006} belong to the former category. \citet{Forman:2004} addresses text classification using binary covariates, treating the $C$-class prediction problem as a set of $C$ one-vs-rest prediction problems, and selecting covariates for each based on a conventional covariate ranking method. This procedure ensures that for each class, covariates are selected that effectively discriminate between that class and the other classes. \citet{Chen:2006} also adopt a one-vs-rest classification approach, performing recursive feature elimination concurrently across all $C$ models. In each iteration, they compute the maximum variable importance values for the $C$ classifiers for each covariate, then remove the covariates for which this maximum is the smallest across the classifiers. The margin difference used by the support vector machines is employed as the VIM in this approach.

Conversely, \citet{Zini:2015} and \citet{Song:2017} target differences among all classes. \citet{Zini:2015} introduces a variant of the group lasso \citep{Yuan:2006} designed to select covariates that effectively distinguish between all classes. \citet{Song:2017} implement an iterative covariate selection process based on the class-specific mean differences calculated for each covariate across all possible pairs of classes. As discussed in \citet{Zini:2015} and in Section~\ref{sec:discussion} of this paper, the strategy of selecting covariates that distinguish well between all classes may not always be appropriate.

\citet{Akinola:2022} provide an overview of multi-class covariate selection methods based on metaheuristic algorithms, which represent a specific category of randomized wrapper algorithms.

Beyond being a VIM, the multi-class VIM also differs in another key aspect from the covariate selection methods for multi-class outcomes discussed above. The fact that it specifically targets covariates exhibiting regions substantially containing observations of specific classes allows for interpretations linked to individual classes. For instance, with an imaginary covariate $X$ ranked high by the multi-class VIM, observations with low $X$ values may typically belong to class $A$, those with high $X$ values to class $B$, and those with very high $X$ values to class $C$.

\section{Multi forests: training and prediction}
\label{sec:mufs}

Like conventional RFs, MuFs are comprised of large numbers of decision trees. Each tree is constructed using a different subsample or bootstrap sample of the observations. This sample is recursively divided by covariate-based splits into progressively smaller subgroups called nodes until a stopping criterion is met. The nodes that are not subdivided further are referred to as end nodes. Unlike conventional RFs, where the splits are exclusively binary---each node dividing into two child nodes---MuFs incorporate multi-way splits, where nodes are divided into more than two child nodes. Specifically, a $K$-way split in a covariate $x_s$ is conducted using $K-1$ split points $I_1 < \dots < I_{K-1}$. Observations with $x_s$ values in the interval $]-\infty, I_1]$ are allocated to the first child node, those in the interval $]I_1, I_2]$ to the second, and so on. Training the trees consists of selecting these covariate-based splits.

Prediction in MuFs, as in conventional RFs, is based on the end nodes in which the observations to be predicted are located after traversing the trees through the covariate-based splits. Predictions can be expressed either as predicted class probabilities or as point predictions of the most likely class of the outcome. Predicted class probabilities are calculated by averaging the frequency distributions of the classes in the end nodes where the observations to be predicted are found, as described by \citet{Malley:2012}. Point predictions can be derived either through the highest predicted class probabilities or through majority voting as in conventional RFs, where the class most frequently appearing in the relevant end nodes is predicted.

The algorithms for training a MuF, prediction, and computing the multi-class VIM and the discriminatory VIM (Section~\ref{sec:vims}) are implemented in the \rcode{diversityForest} R package, currently available on CRAN in version 0.5.0. This package, closely based on the popular RF implementation \rcode{ranger} \citep{Wright:2017}, performs tree construction, prediction, and VIM calculation using a fast C++ implementation.

\subsection{Multi forests as diversity forests}
\label{sec:dfs}

MuFs represent so-called diversity forests (DFs) \citep{Hornung:2022b}. The latter are RFs trained using the DF algorithm that facilitates complex splitting procedures in RFs by drastically reducing the number of candidate splits evaluated during split selection. A DF variant published prior to MuFs is interaction forests \citet{Hornung:2022a}, which model pairwise interactions via bivariable splits and rank bivariate interaction effects between all variable pairs according to their impact on prediction.

The set of candidate splits for each split is found in DFs as follows:

For $m = 1, \dots, nsplits$:
\begin{enumerate}
\item Draw randomly a so-called split problem (described below).
\item From the drawn split problem, select randomly one or several splits, where the probabilities of choosing specific splits can be unequal and interdependent.
\end{enumerate}

The nature of the split problems varies depending on the RF variant. Generally, a split problem encompasses all possible splits with, depending on the RF variant, one or more covariates. In MuFs, split problems consist of the possible binary and multi-way splits that can be performed with the individual covariates. The detailed algorithm for selecting splits in MuFs is more complex and is described in Section~\ref{sec:mufconstr}.

The effectiveness of splits primarily depends on the covariates used rather than the exact positioning of the splits within the covariate values \citep{Hornung:2022b}. This is due to the large variation in predictive information across covariates. Typically, no single split in continuous covariates distinctly outperforms all other possible splits in that covariate, suggesting that the specific choice of split within a covariate is less critical. This variability in split effectiveness across covariates is the key reason for the efficiency of the DF algorithm, as the quality of splits differs more strongly between different split problems than within the same split problem. Consequently, it is sufficient to sample only one or a few candidate splits per split problem to identify effective splits with this approach.

The designation \lq\lq random forests'' for forests trained using the DF algorithm is justified, as in the eponymous article by \citet{Breiman:2001}, this term was used to denote a broader category of forests than the specific method commonly referred to by this term today.

\subsection{Handling of unordered categorical covariates}
\label{sec:categvars}

As previously described, multi-way splitting in MuFs aims to identify regions within the covariate values substantially associated with specific classes. However, this process requires ordered covariate values. Consequently, nominal covariates, that is, categorical covariates without inherent ordering, must be treated differently from other covariates in order to enable their inclusion in multi-way splitting.

This is performed by ordering the categories of nominal covariates prior to training the MuF, treating them in the same manner as metric and ordered categorical covariates. Specifically, an approach outlined in \citet{Coppersmith:1999} is applied. Initially, the frequencies of the respective observations from the different outcome classes are calculated for each category of the categorical covariate. This produces a matrix of dimensions $n_{cat} \times C$, where $n_{cat}$ represents the number of categories and, as before, $C$ represents the number of outcome classes. Principal component analysis (PCA) is then performed on this matrix, with the categories (i.e. the observations in the PCA) weighted according to their frequency of occurrence. The categories are subsequently ordered based on the values of the first principal component derived from the PCA, ensuring that closely positioned categories exhibit similar class distributions. These ordered categories are encoded as $1, \dots, n_{cat}$. This method is available for RFs in the R package \rcode{ranger} when using the option \rcode{respect.unordered.factors="order"} and is discussed and empirically evaluated in \citet{Wright:2019} for conventional RFs.

\subsection{Split selection procedure in the construction of the trees}
\label{sec:mufconstr}

Before we detail the split selection procedure used in the trees to split the nodes, we first offer a simplified representation of the procedure in Algorithm~\ref{alg:multifor} to provide a more accessible overview.

\begin{algorithm}
\caption{Sketch of the split selection procedure in multi forests}
\label{alg:multifor}
\begin{algorithmic}[1]
\State Check whether at least one of three stop criteria is met. If so, stop the procedure.
\State Draw the candidate splits:

\noindent For $s=1, \dots, mtry$:

\noindent Draw a covariate $j_{s}$ and generate one or more multi-way splits in the values of this covariate.
\State Determine the best split:
\begin{enumerate}
\item[\footnotesize 3.1:] Decide randomly whether the split should be multi-way or binary.
\item[\footnotesize 3.2:] If the split is to be multi-way:
\begin{enumerate}
\item[\footnotesize 3.2.1:] Evaluate each multi-way split from the multi-way splits sampled in Step 2 using a split criterion that measures how strongly the classes are represented in the child nodes to which they are assigned.
\item[\footnotesize 3.2.2:] Select the multi-way split that is associated with the best value of the split criterion.
\end{enumerate}
\vspace{0.2cm}
\noindent If the split is to be binary:
\begin{enumerate}
\item[\footnotesize 3.2.1:] Evaluate each split point within the multi-way splits sampled in Step 2 using a split criterion that measures the purity of the two resulting child nodes.
\item[\footnotesize 3.2.2:] Select the binary split associated with the best value of the split criterion.
\end{enumerate}
\end{enumerate}
\end{algorithmic}
\end{algorithm}

We consider four different versions of this split selection procedure, arising from two variants of two components of the algorithm. The first component affects the degree to which the purity of child nodes, with respect to the classes assigned to them, is rewarded. The variant \lq\lq Squared'' rewards this much more than the variant \lq\lq Non-Squared''. The second component pertains to the split criterion used to evaluate the binary splits. The variant \lq\lq Gini'' assesses binary splits as in classic CART trees, whereas the variant \lq\lq Assign Classes'' employs a criterion more similar to that used for the multi-way splits. The latter variant aims to make the values of the discriminatory VIM more comparable in terms of their value ranges to the values of the multi-class VIM.\footnote{Note again that the discriminatory VIM values are calculated based on the split criterion values of the binary splits and the multi-class VIM values are calculated based on the split criterion values of the multi-way splits.} Since these VIMs rely on the split criteria, the choice of versions impacts the calculation of the VIMs as well, which is why these versions will again play a role in Section~\ref{sec:vims}, in which the VIMs are treated.

It is crucial to note that we did not propose these four versions of the split selection procedure as equally viable options, as our conjecture was that these versions tend to give results of varying quality. Instead, they represent alternative possibilities. Based on the simulation study results and the real data analysis, one version will be recommended above the others.

The split selection procedure is described in detail below: 

\begin{enumerate}
\item Stop if at least one of the following conditions is met: a) the current node $l$ consists exclusively of observations of a certain class, that is, it is pure, b) the current node has at most as many observations as the allowed minimum node size $nmin$, which is a tuning parameter, c) none of the available covariates has at least one potential split point, that is, all available covariates are constant in the current node.
\vspace{3mm}
\item Determine the set $\mathcal{C}_l$ of the $c_l$ classes present in the current node $l$.
\vspace{3mm}
\item Sampling of the candidate split variables and the candidate splits.\\[5pt]
Set $m \gets 1$.\\[5pt]
For $s = 1, \dots, mtry$:
\vspace{2mm}
\begin{enumerate}
\item Randomly draw a covariate $x_{j_s}$ from the set of all covariates that have at least one potential split point.
\item Determine the $N$ unique values of $x_{j_s}$ in the current node and sort them. Label the sorted values with $a_1, \dots, a_N$.
\item If $N \leq c_l$:\\
\vspace{-1mm}
\begin{itemize}
\item[] Place the split points $I_{m, 1} < \dots < I_{m, N-1}$ between all neighboring unique $x_{j_s}$ values:
\vspace{2mm}
\[
I_{m, t} = (a_t + a_{t+1})/2, \;\; \text{where} \;\; t = 1, \dots, N-1.
\]
\vspace{0mm}
\item[] Set $m \gets m + 1$.
\end{itemize}
\vspace{3mm}
If $N > c_l$:\\
\vspace{-1mm}
\begin{itemize}
\item[]  Repeat $npervar$ times, where $npervar$ is a tuning parameter:
\vspace{2mm}
\begin{enumerate}
\item Randomly select $c_l - 1$ split points, $I_{m, 1} < \dots < I_{m, c_l - 1}$, from the midpoints between all neighboring unique $x_{j_s}$ values, with the constraint that between all neighboring split points there are at least $\lfloor \frac{N}{2c_l} \rfloor$ unique $x_{j_s}$ values.
\item Set $m \gets m + 1$.
\end{enumerate}
\end{itemize}
\end{enumerate}
\vspace{3mm}
\item Determination of the best split.
\vspace{2mm}
\begin{enumerate}
\item Decide at random whether the split should be multi-way or binary.
\item If the split is to be multi-way:
\vspace{2mm}
\begin{itemize}
\item[] \begin{enumerate}
\item Perform the following for all multi-way splits drawn in Step 3:\\[5pt]
If $N \geq c_l$:\\
\vspace{-1mm}
\begin{itemize}
\item[] In this case, which is the most common case with metric covariates, we assign each class to a different one of the $c_l$ child nodes.\\
More precisely, the classes are assigned to the child nodes in such a way that the following criterion is maximized:
\begin{equation}
\sum_{c \in \mathcal{C}_l} p_{c, \mathcal{I}^{as}_{m,c}}^2.\label{eq:splitcritmuwunw}
\end{equation}
Here $\mathcal{I}^{as}_{m,1}, \dots, \mathcal{I}^{as}_{m,C_l}$ denote the indices of the child nodes to which the classes are assigned, where $\mathcal{I}^{as}_{m,c}$ denotes the child node to which the class $c$ is assigned. Furthermore, $p_{c,\mathcal{I}^{as}_{m,c}}$ denotes the proportion of observations of class $c$ in the $\mathcal{I}^{as}_{m,c}$-th child node. For larger numbers of classes, the number $M!$ of possible assignments of the classes to the child nodes quickly becomes very large. However, finding the optimal allocations $\mathcal{I}^{as}_{m,1}, \dots, \mathcal{I}^{as}_{m,C_l}$ can be carried out computationally efficiently also in such cases using the Hungarian algorithm \citep{Kuhn:1955}.\\
After determining the vector of indices of the child nodes to which the classes are assigned, $\mathcal{I}^{as}_{m,1}, \dots, \mathcal{I}^{as}_{m,C_l}$, the split criterion is calculated as follows:
\begin{equation}
\sum_{c \in \mathcal{C}_l} p_{c, \mathcal{I}^{as}_{m,c}}^2 \frac{n_{l, \mathcal{I}^{as}_{m,c}}}{n_l},\label{eq:splitcritmuw1}
\end{equation}
where $n_{l, \mathcal{I}^{as}_{m,c}}$ denotes the number of observations in the $\mathcal{I}^{as}_{m,c}$-th child node and $n_l$ denotes the number of observations in the current node $l$. In the variant \lq\lq Non-Squared'', both equation \eqref{eq:splitcritmuwunw} and equation \eqref{eq:splitcritmuw1} use the non-squared proportions $p_{c,\mathcal{I}^{as}_{m,c}}$ instead of the squared proportions. The variant \lq\lq Squared'' is correspondingly the variant with squared proportions.
\end{itemize}
\vspace{3mm}
If $N < c_l$:\\
\vspace{-1mm}
\begin{itemize}
\item[] In this case, a similar split criterion is used:
\begin{equation}
\sum_{c \in \mathcal{C}_l} p_{c, \mathcal{I}^{as}_{m,c}}^2 \frac{n_{l, \mathcal{I}^{as}_{m,c}}}{n_l}, \;\; \text{where} \;\; \mathcal{I}^{as}_{m,c} = \argmax_{k \in \{1,N-1\}} p_{c,k}^2. \label{eq:splitcritmuw2}
\end{equation}
The quantities occurring in formula \eqref{eq:splitcritmuw2} are defined in the same way as in formula \eqref{eq:splitcritmuw1}. In the split criterion defined in formula \eqref{eq:splitcritmuw2}, different classes can be assigned to the same child node, unlike in formula \eqref{eq:splitcritmuw1}. This is also unavoidable, as the number of classes is greater than the number of child nodes. If there are several maxima in the proportions $p_{c,k}$ ($k \in \{1, \dots, N-1\}$) in formula \eqref{eq:splitcritmuw2}, $\mathcal{I}^{as}_{m,c}$ is assigned a random index from the indices of the child nodes with maximum proportions. In the variant \lq\lq Non-Squared'', again the non-squared proportions $p_{c, \mathcal{I}^{as}_{m,c}}$ are used in equation \eqref{eq:splitcritmuw2}.
\end{itemize}
\vspace{3mm}
\item Select the (multi-way) split that is associated with the best split criterion value.
\end{enumerate}
\end{itemize}
\vspace{5mm}
If the split is to be binary:
\vspace{2mm}
\begin{itemize}
\item[] Variant \lq\lq Gini'':
\vspace{2mm}
\begin{enumerate}
\item Calculate the Gini split criterion, which is also used in CART trees, for each of the split points sampled in Step 3. That is, in the first step, the Gini split criterion is calculated for the binary split induced by $I_{1,1}$, in the second step it is calculated for the binary split induced by $I_{1,2}$, etc. until all splits generated in Step 3 have been evaluated.
\item Select the split point associated with the best value of the Gini split criterion.
\end{enumerate}
\vspace{2mm}
Variant \lq\lq Assign Classes'':
\vspace{2mm}
\begin{enumerate}
\item As in the variant \lq\lq Gini'', all split points sampled in Step 3 are evaluated. However, the following split criterion is used instead of the Gini criterion:
\begin{equation}
\sum_{c \in \mathcal{C}_l} p_{c, \mathcal{I}^{as}_{m,c}}^2 \frac{n_{l, \mathcal{I}^{as}_{m,c}}}{n_l}, \;\; \text{where} \;\; \mathcal{I}^{as}_{m,c} = \argmax_{k \in \{1,2\}} p_{c,k}^2.\label{eq:splitcritbin}
\end{equation}
The quantities occurring in this formula are defined in the same way as in the formulas \eqref{eq:splitcritmuw1} and \eqref{eq:splitcritmuw2}. If $p_{c,1} = p_{c,2}$, $\mathcal{I}^{as}_{m,c}$ is chosen at random from $\{1,2\}$. In the variant \lq\lq Non-Squared'', again the non-squared proportions $p_{c,k}$ are used in equation \eqref{eq:splitcritbin}.
\item Select the split point that is associated with the best value of the above split criterion.
\end{enumerate}
\end{itemize}
\end{enumerate}
\end{enumerate}

The restriction described in Step 3.(c)i.\ mandates that between any two neighboring split points, there should be at least $\lfloor \frac{N}{2c_l} \rfloor$ unique $x_{j_s}$ values. Note again that in the context of multi-way splits, a different class is assigned to each child node in the calculation of the split criterion and in that of the multi-class VIM. Against this background, the restriction is intended to prevent the child nodes from having substantially disparate sizes. If some child nodes contained few observations, the corresponding classes would be represented by small regions within the covariate's range. These small intervals would hardly allow meaningful or relevant interpretations. This is because knowing that a class exists substantially within a region that contains only a small proportion of the observations provides limited useful insight. Additionally, allowing very small child nodes could lead to unstable VIM values. Node partitions featuring very small child nodes may result in markedly different split criterion values compared to the majority of the split criterion values obtained for the respective covariate. Given that both the multi-class VIM and the discriminatory VIM are calculated based on these split criterion values, such outlying values could adversely affect the stability of the calculated VIM values.

The restriction used in the sampling of the split points mentioned in the last paragraph also affects the binary splits. This is because the split points for the multi-way splits and the binary splits are sampled in the same way. The purpose of this approach is to align the discriminatory VIM values more closely with those of the multi-class VIM.

Unlike the supervised discretization techniques mentioned in Section~\ref{sec:relwork}, MuFs employs a simpler approach for identifying multi-way splits in the values of the randomly sampled covariates. Here, only a small number of $npervar$ candidates of multi-way splits are sampled per covariate, following the DF algorithm. This method might lead to more splits in the trees, as the nodes achieve purity later compared to methods using complex optimization. Consequently, the trees may become more complex and their predictions more varied. This variability can enhance the predictive performance of RFs. As \citet{Breiman:2001} has shown, the predictive performance of RFs benefits from a weak correlation between the predictions of the trees. Given that multi-way splits typically segregate the data rapidly, it is potentially particularly beneficial to avoid optimizing the splits here.

\section{Multi-class VIM and discriminatory VIM}
\label{sec:vims}

As described in the introduction, we propose two distinct types of VIMs: the multi-class VIM and the discriminatory VIM. The multi-class VIM assesses the influence of covariates based on how strongly they are associated with different classes. Specifically, it measures the extent to which regions within the ranges of covariate values are dominated by observations of particular classes. In contrast, the discriminatory VIM, like conventional VIMs, evaluates how effectively covariates can distinguish between the classes. Covariates with high discriminatory VIM values may not necessarily be associated with specific classes, but may only be suitable well for distinguishing between different groups of classes.

The discriminatory VIM is calculated for all covariates, whereas the multi-class VIM is calculated only for continuous covariates and categorical covariates with at least $C$ categories, where $C$ represents the number of outcome classes. The technical reason for this is that the split criterion on which the multi-class VIM is based is calculated differently depending on whether the number of unique values of the covariate is at least as many as the number of classes or fewer. Consequently, the multi-class VIM values would not be equally comparable across covariates with varying numbers of unique values. Additionally, if a covariate has fewer categories than there are classes, it is impossible to separate all $C$ classes well. Therefore, such covariates may arguably have less potential for the multi-class prediction problem.

The multi-class VIM $\text{M\_Cl\_VIM}_s$ and the discriminatory VIM $\text{Disc\_VIM}_s$ for covariate $s$ are calculated as mean values of tree-specific contributions to these measures across all $B$ trees:
\begin{equation}
\text{M\_Cl\_VIM}_s = \frac{1}{B} \sum_{b=1}^B \text{m\_cl\_vim}_{s,b}, \quad \text{Disc\_VIM}_s = \frac{1}{B} \sum_{b=1}^B \text{disc\_vim}_{s,b}
\end{equation}
The tree-specific contributions of the multi-class VIM are defined as follows:
\begin{equation}
\text{m\_cl\_vim}_{s,b} = \left\{
\begin{array}{ll}
\sum_{l \in \mathcal{L}_{s,b}} n_l (\text{oob\_m\_cl\_vim}_l - \text{oob\_m\_cl\_vim\_perm}_l) & \#\{\mathcal{L}_{s,b}\} > 0 \\
0 & \#\{\mathcal{L}_{s,b}\} = 0 \\ \label{eq:mclvimtree}
\end{array}
\right.
\end{equation}
Here $\mathcal{L}_{s,b}$ denotes the set of all nodes in tree $b$ that use covariate $s$ in a multi-way split, where no split in covariate $s$ occurred in the path to this node, neither multi-way nor binary, and are traversed by at least one OOB observation. Additionally, $n_l$ represents the number of (in-bag) observations in node $l$. Moreover, $\text{oob\_m\_cl\_vim}_l $ is the value of the multi-way split criterion calculated on the OOB data, where only those OOB observations are used that pass through node $l$:
\begin{equation}
\text{oob\_m\_cl\_vim}_l = \sum_{c \in C_l} p_{OOB, c, \mathcal{I}^{as}_c}^2, \label{eq:oobmclvim}
\end{equation}
where $p_{OOB, c, \mathcal{I}^{as}_c}$ is the proportion of OOB observations in child node $\mathcal{I}^{as}_c$ of node $l$ that are from class $c$ and $\mathcal{I}^{as}_c$ is the child node to which class $c$ was assigned during the construction of the tree, that is, using the in-bag data. In the variant \lq\lq Non-Squared'' (see Section~\ref{sec:mufconstr}), the non-squared proportions $p_{OOB, c, \mathcal{I}^{as}_c}$ are used in equation \eqref{eq:oobmclvim} accordingly. As mentioned in the introduction, the rationale for using only OOB data in the calculation of the VIM is to avoid biases caused by overfitting similar to those of the (uncorrected) Gini importance.

To calculate $\text{oob\_m\_cl\_vim\_perm}_l$, the values of the covariate $s$ in the OOB observations that pass node $l$ are randomly permuted and then processed in the same manner as for the calculation of $\text{oob\_m\_cl\_vim}_l$.

Subtracting the split criterion calculated on the OOB data after permutation of the covariate serves two purposes. First, it makes the multi-class VIM values of non-informative covariates fluctuate around zero, facilitating the assessment of which covariates are likely to have an effect and which are not. Second, this subtraction helps correct for potential bias. Among others, \citet{Loecher:2020} demonstrated that VIMs calculated based on OOB data can also be biased. The principle of calculating a measure on the OOB data and then subtracting the corresponding value after permutation of the covariate is also applied in the calculation of the permutation VIM. However, with the latter, this procedure results in correlated covariates receiving variable importance values that are too high. This is because the permutation of covariates that are strongly correlated with others leads to extrapolations into areas of the covariate space with low data density, where the trees provide poor predictions \citep{Hooker:2021}. This issue does not arise with our approach, since the permutation only affects the discriminatory capacity of the split at the node where it is performed and not also that of splits in other covariates at other locations in the trees.

The rationale for considering only those nodes in the split criterion for which there has not yet been a split in the covariate, either multi-way or binary, in the path to the node is that this prevents bias in the VIM. After the first multi-way split has been performed in a covariate, this covariate naturally has hardly any influence on the outcome in all subsequent nodes below the split node. Similarly, after the first binary split in a covariate, it is likely that this covariate will have a strongly attenuated and modified influence in all subsequent nodes. Therefore, the multi-way split criterion values for this covariate in these nodes do not reflect the actual effect strength of the covariate.

Lastly, the reasons why the tree-specific VIM contributions in formula \eqref{eq:mclvimtree} are multiplied by the node size $n_l$ are twofold. On one hand, this multiplication accounts for the fact that splits in larger nodes are more influential. On the other hand, it reflects that covariates with stronger effects tend to be used for splitting earlier than those with weaker effects.

The calculation of $\text{disc\_vim}_{s,b}$ is carried out as follows:
\begin{equation}
\text{disc\_vim}_{s,b} = \left\{
\begin{array}{ll}
\sum_{l \in \tilde{\mathcal{L}}_{s,b}} n_l (\text{oob\_disc\_vim}_l - \text{oob\_disc\_vim\_perm}_l) & \#\{\tilde{\mathcal{L}}_{s,b}\} > 0 \\
0 & \#\{\tilde{\mathcal{L}}_{s,b}\} = 0 \\ \label{eq:discvimtree}
\end{array}
\right.
\end{equation}
Here, $\tilde{\mathcal{L}}_{s,b}$ represents the set of all nodes in tree $b$ that use covariate $s$ in a binary split, where in the path to this node there was no previous split in covariate $s$, neither multi-way nor binary, and are traversed by at least one OOB observation. As in formula \eqref{eq:oobmclvim}, $n_l$ denotes the number of observations in node $l$. In the variant \lq\lq Assign Classes'', the quantities $\text{oob\_disc\_vim}_l$ and $\text{oob\_disc\_vim\_perm}_l$ are calculated analogously as in the case of the multi-class VIM. Here, in the variant \lq\lq Non-Squared'', the non-squared proportions $p_{OOB, c, \mathcal{I}^{as}_c}$ are used accordingly. In the variant \lq\lq Gini'', the Gini index, calculated exclusively using the corresponding OOB data, is used in the calculation of $\text{oob\_disc\_vim}_l$ and $\text{oob\_disc\_vim\_perm}_l$.

The reasons for the different components of the discriminatory VIM are analoguous to those for the multi-class VIM.

The discriminatory VIM serves the same purpose as conventional VIMs such as the permutation VIM. However, the procedure of calculating the discriminatory VIM is very similar to that of the multi-class VIM. Consequently, the values of the discriminatory VIM can be expected to be more directly comparable with those of the multi-class VIM than VIMs calculated by a completely different method. By comparing the covariate-specific multi-class and discriminatory VIM values, it may become more convenient to identify which covariates are specifically linked with individual classes and which can distinguish effectively merely between groups of classes.

It is essential that the multi-class VIM is based on the split criterion used for the multi-way splits. This approach ensures that the VIM measures the importance of the covariates specifically in terms of their association with individual classes as opposed to groups of classes. This presents a different purpose compared to in the case of the Gini importance, which also uses the split criterion in the computation of the VIM values. The Gini importance assesses the general influence of the covariates on the prediction, without focusing on specific types of this influence.

\section{Simulation study: Comparison of multi-class VIM and discriminatory VIM with conventional VIMs}
\label{sec:simstudy}

The motivation behind developing the multi-class VIM was to create a VIM for data with multi-class outcomes, capable of identifying covariates specifically associated with one or more classes. More specifically, it should identify what we defined in the introduction as \lq\lq class-associated covariates'' -- covariates with value regions where observations of certain classes are substantially found.

To identify class-associated covariates, they must be ranked higher by the multi-class VIM than other influential covariates that only distinguish well between groups of classes. In this simulation study, we investigate the extent to which this is the case. We also evaluate how conventional VIMs perform in this respect. If conventional VIMs also consistently rank the class-associated covariates highest, the multi-class VIM would not offer added value compared to conventional VIMs.

The discriminatory VIM proposed alongside the multi-class VIM ranks covariates similarly to conventional VIMs. It assesses how well covariates separate observations of different classes. However, it does not consider how specifically covariates are associated with individual classes. In this simulation study, we also investigate the extent to which the discriminatory VIM provides results comparable to conventional VIMs. Furthermore, we examine whether comparing the multi-class VIM values with the discriminatory VIM values provides a clearer picture. This comparison may clarify even better which covariates are class-associated. It may also indicate which covariates are only suitable for distinguishing between groups of classes.

All R code necessary to reproduce the simulation study and the real data analysis (Section~\ref{sec:rdanalysis}), along with the datasets used, is available on Github (\url{https://github.com/RomanHornung/MultiForests_code_and_data}).

\subsection{Study design}

\subsubsection{Compared methods}
\label{sec:comp_methods}

The following VIMs were compared: multi-class VIM, discriminatory VIM, permutation VIM, and corrected Gini importance \citep{Nembrini:2018}. As described in Section~\ref{sec:mufconstr}, we consider four different versions of the MuF algorithm, also affecting the calculation of the multi-class and discriminatory VIMs. These variants are identified as follows in the presentation of the results: \rcode{wsquared\_wgini} for the version with squared proportions $p_{c, \mathcal{I}^{as}_{m,c}}^2$ and split criterion \lq\lq Gini'' for the binary splits, \rcode{wosquared\_wgini} for the version without squaring the proportions $p_{c, \mathcal{I}^{as}_{m,c}}$ and split criterion \lq\lq Gini'' for the binary splits, \rcode{wsquared\_wogini} for the version with squared proportions $p_{c, \mathcal{I}^{as}_{m,c}}^2$ and split criterion \lq\lq Assign Class'' for the binary splits, and \rcode{wosquared\_wogini} for the version without squaring the proportions $p_{c, \mathcal{I}^{as}_{m,c}}$ and split criterion \lq\lq Assign Class'' for the binary splits. The permutation VIM and the corrected Gini importance are denoted by \rcode{perm} and \rcode{gini\_corr}, respectively.

Per MuF, $ntree=5000$ trees were trained with $npervar=5$ randomly sampled multi-way splits per sampled covariate. These values were chosen based on an informal analysis of some datasets used in Section~\ref{sec:rdanalysis} (results not shown). This analysis suggested that the value $npervar=5$ provides sufficiently stable VIMs, with highly correlated VIM values between two forests trained on the same data. Furthermore, subsampling instead of bootstrap was used to draw the observations for each tree. Here, a proportion of $prop=0.7$ of the data was subsampled in each case. This choice was based on real data analyses using many datasets presented in \citet{Probst:2019}, where subsampling with this proportion provided the best prediction results on average with conventional RFs. The parameter $mtry$, which represents the number of candidate covariates evaluated at each split in the tree, was set to $\lfloor \sqrt{p} \rfloor$, where $p$ is the total number of covariates in the dataset. This is the most common default choice for this parameter in RFs.

For the RFs trained to compute permutation VIM and corrected Gini importance, $ntree=5000$, $mtry = \lfloor \sqrt{p} \rfloor$, and subsampling with $prop=0.7$ were also used. This ensured that the comparison of the VIMs is not confounded by different method configurations.

\subsubsection{Data generating processes}

We simulated data with normally distributed covariates and multi-class outcomes, considering three settings for the number of outcome classes: $C = 4, 6, 10$. The classes were balanced, that is, each represented by approximately the same number of observations. For each value of $C$, the following sample sizes were considered: $n = 100, 500, 1000, 2000$. For each combination of $C$ and $n$, 500 datasets were simulated and analyzed using the different variable importance methods.

Each dataset contained 50 standard-normally distributed and uncorrelated noise covariates without mean differences between classes, denoted as $X_{no}$. Informative covariates were divided into covariates with mean differences between two ($X_{two\_gr}$) or three ($X_{thr\_gr}$) groups of classes, and class-associated covariates where one or more classes had different means than all other classes ($X_{cl\_as\_1}$, $X_{cl\_as\_2}$, $X_{cl\_as\_3}$, $X_{cl\_as\_4}$). For each type of informative covariate considered, that is, for $X_{two\_gr}$, $X_{thr\_gr}$, $X_{cl\_as\_1}$ etc., three covariates were generated per dataset. Depending on the number of outcome classes, the sets of informative covariates differed slightly. All covariates had a variance of 1 within classes.

Table~\ref{tab:simparams} shows which covariates were present in each setting and their class-specific mean values. The class-specific distributions of the covariates are shown in Figure~S1 (supplementary material). The class-specific mean values of the informative covariates were chosen such that the differences between neighboring classes, with respect to the covariate values, were 0.75 or 1. These differences were classified as moderate and strong differences, respectively, in the simulation study by \citet{Janitza:2013}. An exception was made for $X_{tw\_gr}$, where the differences were set at a larger value of 1.5. This was because only two groups of classes differed for this covariate, with no mean differences within these groups, necessitating a greater difference between the two groups to obtain a signal strength similar to that of other informative covariates.

\begin{table}[h]
\caption{Means of the covariates in the simulation study.}\label{tab:simparams}
\begin{tabular}{l |c c c c c c c}
\hline
& \multicolumn{7}{c}{$C = 4$} \\
\hline
& $X_{no}$ & $X_{tw\_gr}$ & $X_{thr\_gr}$ & $X_{cl\_as\_1}$ & $X_{cl\_as\_2}$ & $X_{cl\_as\_3}$ & $X_{cl\_as\_4}$ \\
\hline
$c=1$ & 0 & 0 & & 0 & 0 & 0 & \\
$c=2$ & 0 & 0 & & 0 & 0 & 0.75 & \\
$c=3$ & 0 & 1.5 & & 0 & 1 & 1.5 & \\
$c=4$ & 0 & 1.5 & & 1 & 2 & 2.25 & \\
\hline
& \multicolumn{7}{c}{$C = 6$} \\
\hline
$c=1$ & 0 & 0 & 0 & 0 & 0 & 0 & \\
$c=2$ & 0 & 0 & 0 & 0 & 0 & 0 & \\
$c=3$ & 0 & 0 & 1 & 0 & 0 & 0 & \\
$c=4$ & 0 & 1.5 & 1 & 0 & 0 & 0.75 & \\
$c=5$ & 0 & 1.5 & 2 & 0 & 1 & 1.5 & \\
$c=6$ & 0 & 1.5 & 2 & 1 & 2 & 2.25 & \\
\hline
& \multicolumn{7}{c}{$C = 10$} \\
\hline
$c=1$ & 0 & 0 & 0 & 0 & 0 & 0 & 0 \\
$c=2$ & 0 & 0 & 0 & 0 & 0 & 0 & 0 \\
$c=3$ & 0 & 0 & 0 & 0 & 0 & 0 & 0 \\
$c=4$ & 0 & 0 & 0 & 0 & 0 & 0 & 0 \\
$c=5$ & 0 & 0 & 1 & 0 & 0 & 0 & 0.75 \\
$c=6$ & 0 & 1.5 & 1 & 0 & 0 & 0 & 0.75 \\
$c=7$ & 0 & 1.5 & 1 & 0 & 0 & 0.75 & 1.5 \\
$c=8$ & 0 & 1.5 & 2 & 0 & 0 & 0.75 & 1.5 \\
$c=9$ & 0 & 1.5 & 2 & 0 & 1 & 1.5 & 2.25 \\
$c=10$ & 0 & 1.5 & 2 & 1 & 2 & 2.25 & 3 \\
\hline
\end{tabular}
\end{table}

The mean differences for $X_{cl\_as\_3}$ and $X_{cl\_as\_4}$ were chosen smaller than for $X_{cl\_as\_1}$ and $X_{cl\_as\_2}$ because more classes differ for the former. The more classes differ, the larger the VIM values of the corresponding covariates become. If the same mean differences had been used, the VIM values for $X_{cl\_as\_3}$ and $X_{cl\_as\_4}$ would have been very large, incomparably larger than those of $X_{cl\_as\_1}$ and $X_{cl\_as\_2}$.

Table~\ref{tab:simparams} shows that for $C = 10$, some pairs of classes have the same mean values for $X_{cl\_as\_3}$ and $X_{cl\_as\_4}$. This is because, in scenarios with large numbers of classes, it is of interest to also identify covariates that contain regions specific to small subsets of classes, rather than to individual classes exclusively. In applications with many classes, it is unrealistic to have covariates that can distinguish between a large proportion of the classes. Moreover, if such covariates are present, they are expected to have a strong signal, receiving very large variable importance values even from conventional VIMs.

\subsubsection{Evaluation}

For evaluation, we used an approach similar to that of \citet{Janitza:2013}, which is based on the Area Under the ROC Curve (AUC). Typically, the AUC is used in the context of diagnostic tests where the test results are continuous risk scores. In this context, the AUC represents the probability that the test assigns a higher risk score to a randomly selected diseased patient than to a randomly selected healthy patient.

We used the AUC to measure how strongly the VIMs under consideration assign different importance values to different types of covariates. Specifically, in each case we examined two different types of covariates. For example, we compared the covariates $X_{two\_gr}$, which have mean differences only between two groups of classes, and the noise covariates $X_{no}$ without mean differences. Here, the corresponding AUC would represent the probability that a randomly selected $X_{two\_gr}$ covariate receives a higher VIM value than a randomly selected noise covariate. Another example is the comparison between the covariates $X_{cl\_as\_2}$, where two classes have different mean values than all other classes, and the covariates $X_{two\_gr}$. Here, the AUC would represent the probability that a randomly selected $X_{cl\_as\_2}$ covariate receives a higher VIM value than a randomly selected $X_{two\_gr}$ covariate. Such interpretations are more useful than comparing raw VIM values. This is because the AUC values quantify the extent to which the different VIMs assign higher values to the respective covariates of interest than to other covariates.

Each AUC value was calculated as the mean of 500 AUC values, each derived from one of the 500 simulated datasets for each setting. Each AUC value for the individual datasets was calculated using only the VIM values obtained for the two types of covariates considered. For example, when comparing the covariate types $X_{cl\_as\_2}$ and $X_{two\_gr}$, only six VIM values could be used per dataset, as each dataset contained only three covariates of each type. Consequently, the individual AUC values calculated from the datasets have a high variance. However, after averaging over the 500 datasets, the AUC values had an acceptably low variance. This was verified by calculating a 95\% confidence interval for each of the AUC values obtained after averaging. In the calculation of these confidence intervals we utilized the fact that means of independent numerical values are asymptotically normally distributed, even if the individual values are not normally distributed.

As described at the beginning of this section, a further aim of the simulation study was to investigate the comparison of the multi-class VIM values with the discriminatory VIM values. This comparison may allow an even clearer picture of which covariates are class-associated and which are merely suited to distinguish between groups of classes. To investigate this, in addition to the raw multi-class VIM values, we also considered the differences between the multi-class VIM values and the corresponding discriminatory VIM values in place of the raw VIM values in the AUC analysis.

\subsection{Results}

\begin{figure}[ht!]
\begin{center}
\includegraphics[width = \textwidth]{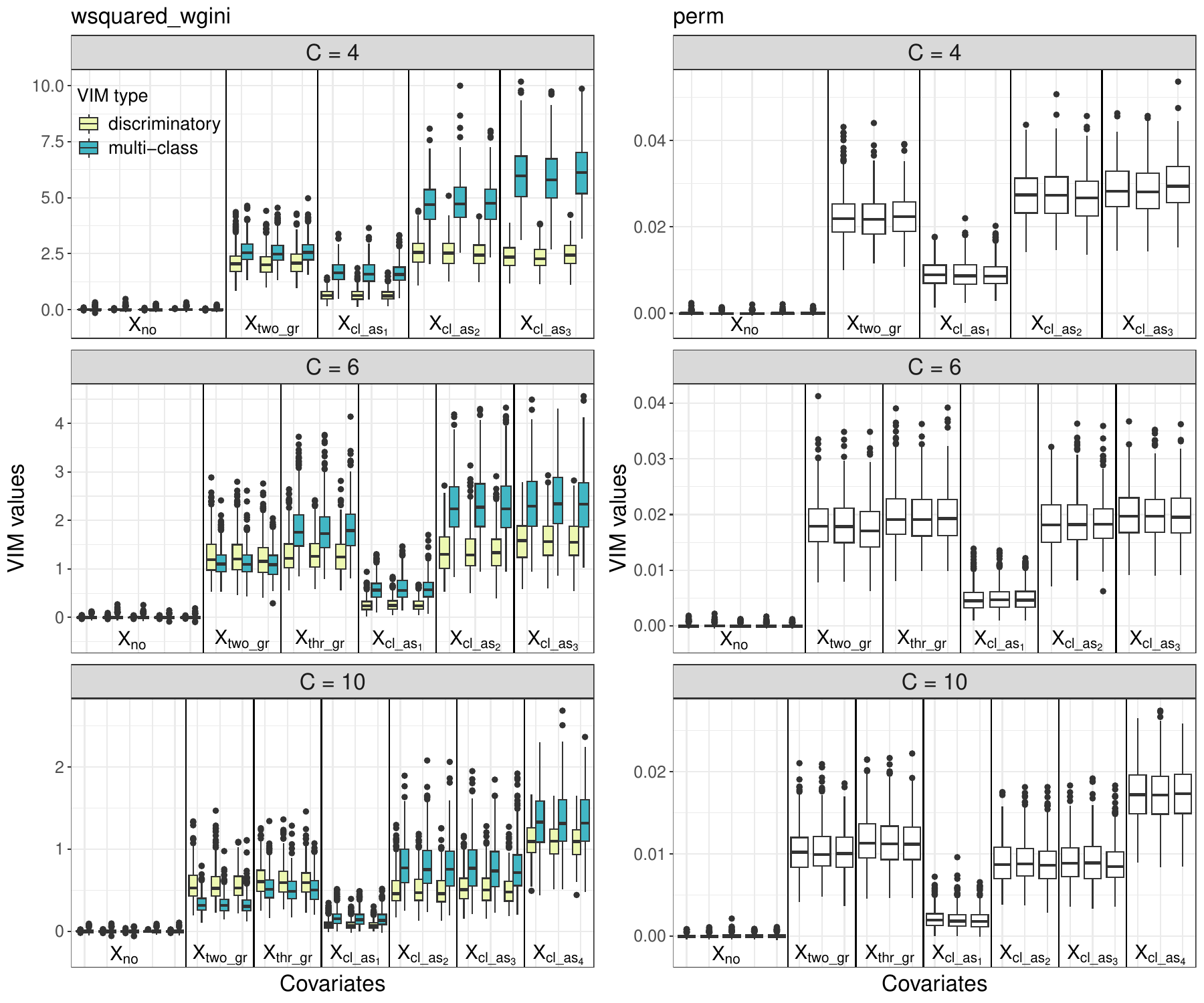}
\end{center}
\caption{VIM values obtained for \rcode{wsquared\_wgini} and the permutation VIM (\rcode{perm}) obtained for all simulated datasets with $n=500$. For visual clarity, the VIM values of only five of the 50 noise covariates are shown.}
\label{fig:vimvalues}
\end{figure}

\myheading{Raw VIM values}
\begin{sloppypar}
Figure~\ref{fig:vimvalues} shows the raw VIM values of the multi-class VIM and the discriminatory VIM for the \rcode{wsquared\_wgini} version and those of \rcode{perm} for all datasets with $n=500$. We focus on the \rcode{wsquared\_wgini} version because, as we will show later, this version will be the recommended one. This recommendation will be based on the results described in this section and in Section~\ref{sec:rdanalysis}. Here, this version performed best in ranking class-associated covariates high and in terms of prediction, respectively. Figures~S2 to S13 (supplementary material) show the multi-class and discriminatory VIM values for all four versions for all four $n$ values considered.
\end{sloppypar}

The discriminatory VIM in Figure~\ref{fig:vimvalues} behaves reassuringly similar to the permutation VIM. Both VIMs assigned predominantly high VIM values both to the covariates that distinguish between groups of classes ($X_{two\_gr}$, $X_{thr\_gr}$) and to the class-associated covariates. One exception was $X_{cl\_as\_1}$, which was assigned the lowest values among the informative covariates for all VIMs. However, $X_{cl\_as\_1}$ still had consistently higher VIM values than the noise covariates. The overlap of VIM values between informative and non-informative covariates would have been greater with smaller effect sizes. However, as shown by the class-specific distributions of the covariates in Figure~S1 (supplementary material), the effect sizes are not unrealistically large. It is expected that the VIMs reliably assign larger values to covariates with moderate to strong signals than to pure noise covariates. We did not focus on the general ability of the VIMs to distinguish between informative and non-informative covariates. Instead, our attention was on the degree to which the multi-class VIM assigns higher values to covariates specifically associated with individual classes, as opposed to those that merely distinguish well between groups of classes. Additionally, we explored how this VIM differs from conventional VIMs and the discriminatory VIM in this regard.

Figure~\ref{fig:vimvalues} confirms that, unlike the discriminatory VIM and \rcode{perm}, the multi-class VIM tends to assign higher values to class-associated covariates. This is generally true with the exception of $X_{cl\_as\_1}$. In comparison, it assigns lower values to covariates that only distinguish well between groups of classes. In the descriptions below, we will quantify these differences and also include the other three versions of the MuF algorithm and the corrected Gini importance in our comparison.

\myheading{Comparability between multi-class VIM values and discriminatory VIM values for the different MuF versions}
The motivation behind considering the variant \lq\lq Assign Class'' was that the split criterion used in this case for the binary splits is more similar to the split criterion used for the multi-way splits than the Gini split criterion. Consequently, the resulting discriminatory VIM values could have more similar value ranges to those of the multi-class VIM than when using the Gini split criterion (variant \lq\lq Gini''). However, this is not reflected in the results of our simulation.

A comparison of the discriminatory and multi-class VIM values across the different MuF versions shows that the discriminatory VIM values differ strongly between the different versions (Figures~S2 to S13 in the supplementary material). This difference is especially notable between the variants \lq\lq Assign Class'' and \lq\lq Gini''. However, the discriminatory VIM values for the variant \lq\lq Assign Class'' are not more similar to the corresponding multi-class VIM values than for the variant \lq\lq Gini''.

Upon reflection, this is not surprising. The split criterion used for the \lq\lq Assign Class'' variant is only superficially similar to the split criterion used for multi-way splitting. An important difference is that with binary splitting, more than one class is assigned to each sub-node when calculating the split criterion. In contrast, with multi-way splitting, each sub-node is assigned a unique class, except when the number of unique covariate values is less than the number of outcome classes. An exception where the discriminatory VIM values are much more similar to the multi-class VIM values for the variant \lq\lq Assign Class'' than for the variant \lq\lq Gini'' are the noise covariates (Figure~S14 in the supplementary material). In the case of these covariates, however, this comparability is not relevant. This is because interpretations regarding the nature of the influence of noise covariates are naturally not meaningful.

In summary, the presumed advantage that the variant \lq\lq Assign Class'' would lead to discriminatory VIM values more comparable to those of the multi-class VIM is not confirmed by the results of the simulation study.

\myheading{AUC analysis: Influential versus noise covariates}
The mean AUC values with confidence intervals for the comparison of the influential covariates with the noise covariates can be found in the Tables~S1 to S3 (supplementary material). For all VIMs, the influential covariates almost always had higher VIM values than the noise covariates. Only in the case of the smallest number of cases $n=100$ and almost exclusively for $X_{cl\_as\_1}$ were the AUC values noticeably smaller than 1. For $C=4$, the values obtained for $X_{cl\_as\_1}$ were around 0.95. For $C=6$, they were between 0.8 and 0.9, and for $C=10$, they were between 0.65 and 0.8.

In the case of the discriminatory VIM, the AUC values for \rcode{wsquared\_wgini} and \rcode{wosquared\_wgini} were smaller than those for \rcode{wsquared\_wogini} and \rcode{wosquared\_wogini}. For the latter two, the AUC values were comparable to those for the conventional VIMs \rcode{perm} and \rcode{gini\_corr}. Nevertheless, in general, the discriminatory VIM associated with the variant \lq\lq Gini'' does not appear to be notably disadvantaged compared to the discriminatory VIM associated with the variant \lq\lq Assign Class''. For larger numbers of cases or covariates with stronger influences than $X_{cl\_as\_1}$, both variants behaved similarly. They provided consistently larger VIM values for the influential covariates than for the noise covariates.

The AUC values obtained for $X_{cl\_as\_1}$ and $n=100$ with the multi-class VIM were consistently slightly smaller than the AUC values obtained with the conventional VIMs but differed only slightly among the four different variants. The confidence intervals are narrow across settings, only in one case wider than 0.03.

\begin{figure}[ht!]
\begin{center}
\includegraphics[width = \textwidth]{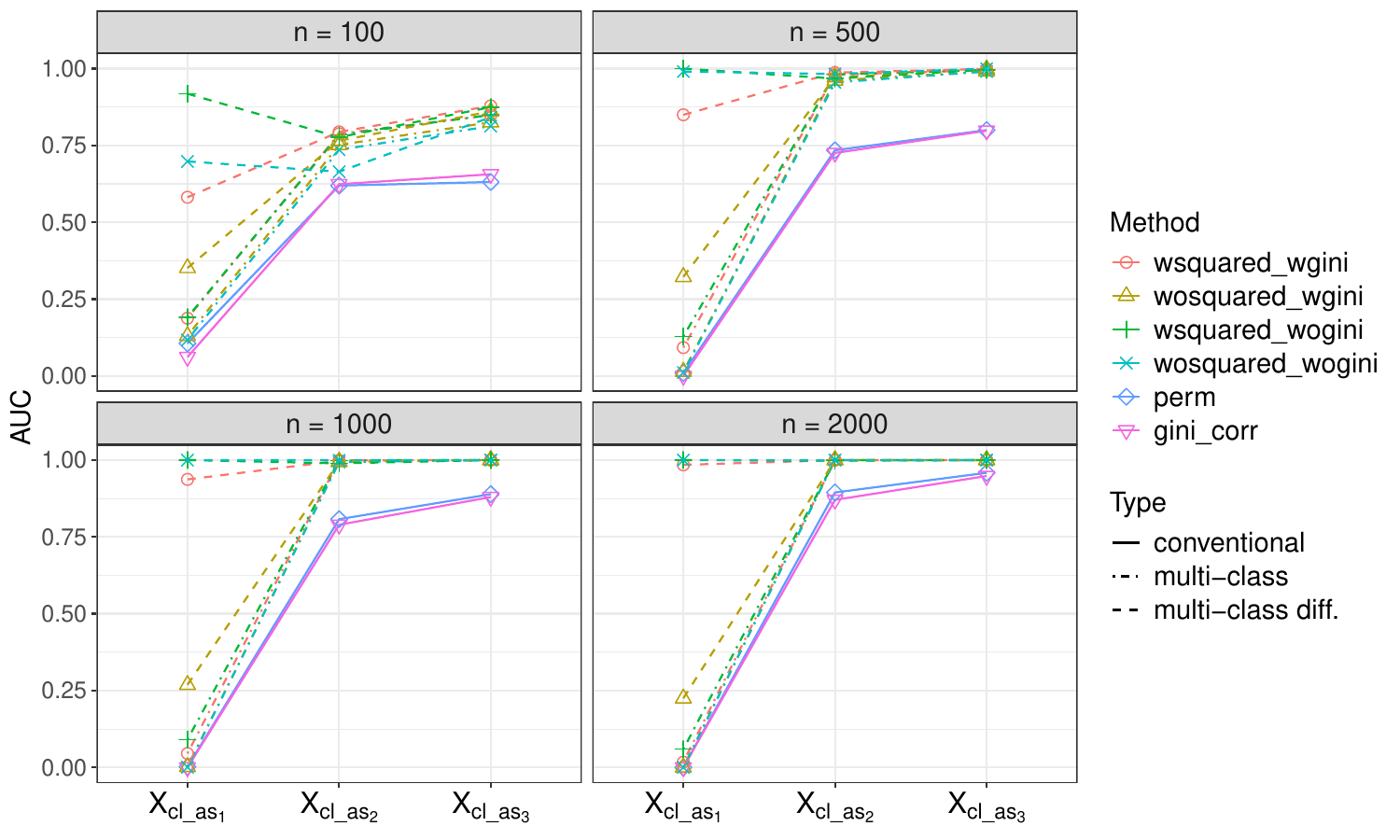}
\end{center}
\caption{Mean AUC values per considered sample size and method for $C=4$. The line types distinguish the different VIM types, where \rcode{conventional} corresponds to conventional VIMs, \rcode{multi-class} to multi-class VIMs, and \rcode{multi-class diff.} to the differences between the multi-class VIM values and the corresponding discriminatory VIM values.}
\label{fig:AUC_C4}
\end{figure}

\myheading{AUC analysis: Class-associated covariates versus $X_{two\_gr}$ -- $C = 4$}
The mean AUC values for the comparison of the VIM values of the class-associated covariates with those of the $X_{two\_gr}$ covariates for the setting $C=4$ are shown in Figure~\ref{fig:AUC_C4}. The AUC values of the multi-class VIMs are generally larger than those for the conventional VIMs.

\begin{sloppypar}
There are only differences between the various multi-class VIM versions for $X_{cl\_as\_2}$ and $X_{cl\_as\_3}$ in the case of the smaller sample sizes. Here, \rcode{wosquared\_wogini} performed the worst and \rcode{wosquared\_wgini} the second worst. For larger numbers of cases, the AUC values for all multi-class VIMs are close to or equal to one. This indicates that the multi-class VIM values of the $X_{cl\_as\_2}$ and $X_{cl\_as\_3}$ covariates were almost always greater than those of the $X_{two\_gr}$ covariates. For larger sample sizes, the AUC values of the conventional VIMs also approached one but remained further behind those of the multi-class VIM versions. The multi-class VIM values obtained for $X_{cl\_as\_1}$ were smaller than the multi-class VIM values obtained for $X_{two\_gr}$ in the vast majority of cases. However, except in the case of \rcode{wosquared\_wgini}, the AUC values of $X_{cl\_as\_1}$ were very high for all multi-class VIM versions when the discriminatory VIM values were subtracted.
\end{sloppypar}

The reason for this is that the drop from $X_{two\_gr}$ to $X_{cl\_as\_1}$ was lower for these multi-class VIM versions than for the corresponding discriminatory VIM versions. Thus, a comparison between the multi-class VIM and the discriminatory VIM would have offered added value here. Only such a comparison could have indicated that $X_{cl\_as\_1}$ is a class-associated covariate.

\begin{figure}[ht!]
\begin{center}
\includegraphics[width = \textwidth]{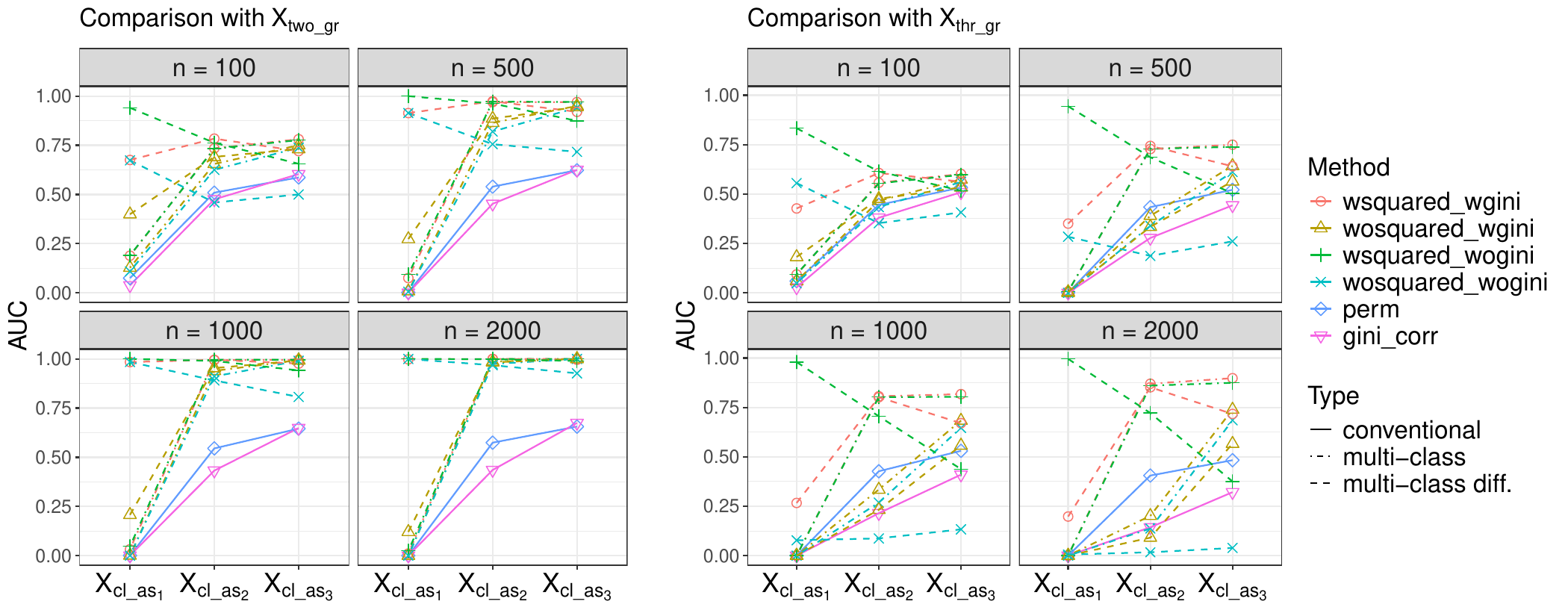}
\end{center}
\caption{Mean AUC values per considered sample size and method for $C=6$. The line types distinguish the different VIM types, where \rcode{conventional} corresponds to conventional VIMs, \rcode{multi-class} to multi-class VIMs, and \rcode{multi-class diff.} to the differences between the multi-class VIM values and the corresponding discriminatory VIM values.}
\label{fig:AUC_C6}
\end{figure}

\myheading{AUC analysis: Class-associated covariates versus $X_{two\_gr}$ and $X_{thr\_gr}$ -- $C = 6$}
Figure~\ref{fig:AUC_C6} shows the comparison of the VIM values between the class-associated covariates and the $X_{two\_gr}$ and $X_{thr\_gr}$ covariates for the setting $C=6$. As in the setting $C=4$, the multi-class VIM values of the covariates $X_{cl\_as\_2}$ and $X_{cl\_as\_3}$ are almost always greater than those of the $X_{two\_gr}$ values. This is not observed for the conventional VIMs. Here, the AUC values for these covariates are always approximately 0.5, regardless of the number of cases considered. This illustrates that the conventional VIMs are conceptually not suitable for distinguishing the class-associated covariates from the $X_{two\_gr}$ covariates.

As in the setting $C=4$, the versions \rcode{wosquared\_wogini} and \rcode{wosquared\_wgini} perform worst. The versions \rcode{wsquared\_wgini} and \rcode{wsquared\_wogini} perform very similarly, with \rcode{wsquared\_wgini} being associated with slightly higher AUC values.

The differences between the multi-class VIM values of the class-associated covariates and the $X_{thr\_gr}$ covariates are much smaller than those of the former and the $X_{two\_gr}$ covariates. Only the \rcode{wsquared\_wgini} and \rcode{wsquared\_wogini} versions provided larger AUC values than the conventional VIMs. These values were around 0.8, indicating that the multi-class VIM values of the class-associated covariates were larger than those of the $X_{thr\_gr}$ covariates in most cases. 

For $X_{cl\_as\_2}$ and especially $X_{cl\_as\_3}$, the AUC values of the multi-class VIM values after subtracting the discriminatory VIM values are worse than those of the raw multi-class VIM values. The cause of this is particularly evident in the case of $X_{cl\_as\_3}$ by examining the raw VIM values (Figures~S6 to S9 in the supplementary material). Here, not only are the multi-class VIM values for $X_{cl\_as\_3}$ higher than those for the other covariates, but the same trend holds for the discriminatory VIM values, albeit to a lesser extent. Therefore, subtracting the discriminatory VIM values results in differences for $X_{cl\_as\_3}$ that are less distinct from those for $X_{thr\_gr}$ than when using the raw multi-class VIM values.

As also seen for $C = 4$, in the comparison with $X_{two\_gr}$, the AUC values for $X_{cl\_as\_1}$ are much larger in many cases after subtracting the discriminatory VIM values. It is noteworthy that in the comparison with $X_{thr\_gr}$, only the AUC value of \rcode{wsquared\_wogini} is high when the corresponding discriminatory VIM values are subtracted. This is because, for \rcode{wsquared\_wogini}, the drop in the discriminatory VIM values from $X_{thr\_gr}$ to $X_{cl\_as\_1}$ is much greater than the drop in the multi-class VIM values. This pattern is not observed for the other MuF versions.

\begin{figure}[ht!]
\begin{center}
\includegraphics[width = \textwidth]{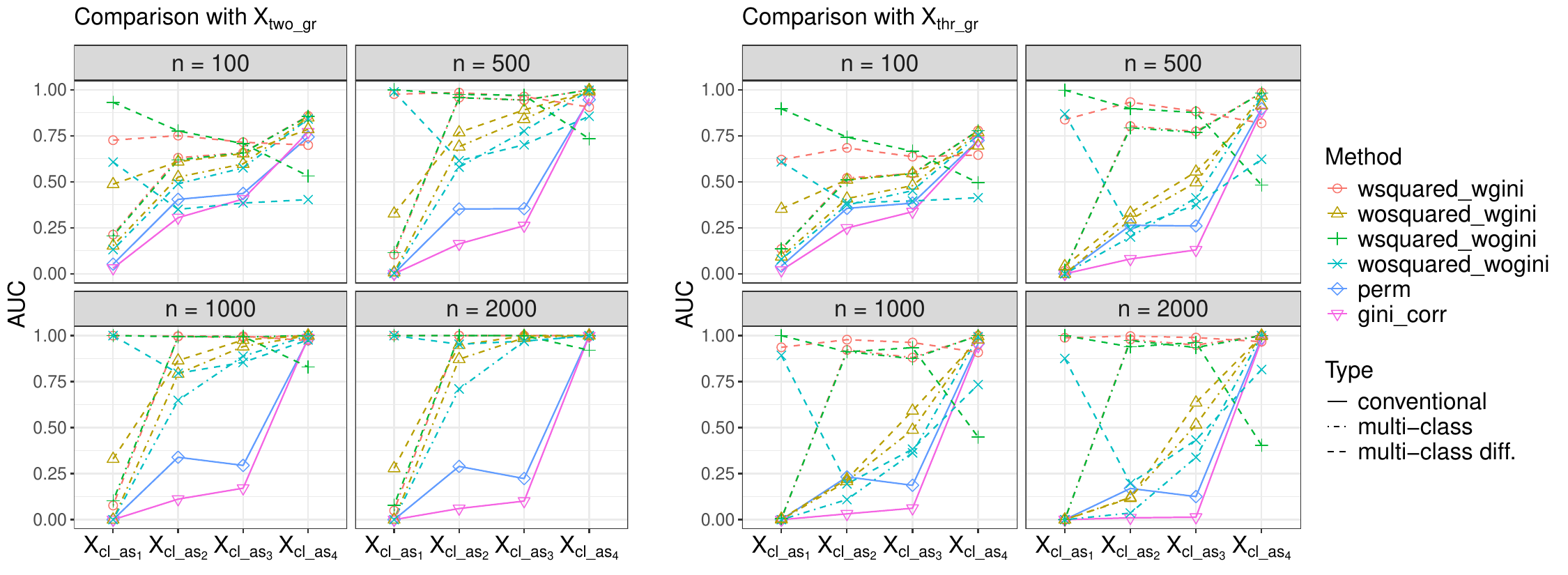}
\end{center}
\caption{Mean AUC values per considered sample size and method for $C=10$. The line types distinguish the different VIM types, where \rcode{conventional} corresponds to conventional VIMs, \rcode{multi-class} to multi-class VIMs and \rcode{multi-class diff.} to the differences between the multi-class VIM values and the corresponding discriminatory VIM values.}
\label{fig:AUC_C10}
\end{figure}

\myheading{AUC analysis: Class-associated covariates versus $X_{two\_gr}$ and $X_{thr\_gr}$ -- $C = 10$}
The AUC values obtained for the $C=10$ setting are shown in Figure~\ref{fig:AUC_C10}. Here, \rcode{wsquared\_wgini} and \rcode{wsquared\_wogini} delivered the best results by a wide margin. The differences between the AUC values achieved for these two methods are very small. It is particularly striking that only these two methods assigned larger multi-class VIM values to the covariates $X_{cl\_as\_2}$ and $X_{cl\_as\_3}$ than to the $X_{thr\_gr}$ covariates. With the exception of $X_{cl\_as\_4}$, the conventional VIMs largely assigned smaller values to the class-associated covariates than to the $X_{two\_gr}$ and $X_{thr\_gr}$ covariates. The covariate $X_{cl\_as\_4}$ received the highest values among all covariates for all VIMs, including the conventional VIMs. This suggests that the influence of $X_{cl\_as\_4}$ is so strong that it is also prioritized by the conventional VIMs, even though other covariates with strong influences but no class-specific association are present.

This demonstrates that conventional VIMs do not automatically rank class-associated covariates lower than covariates without class-specific influence. However, the multi-class VIM helps to specifically select class-associated covariates, as it ranks covariates without class-specific influence lower. After subtracting the discriminatory VIM values, the AUC values for $X_{cl\_as\_4}$ for two MuF versions (\rcode{wsquared\_wogini}, \rcode{wosquared\_wogini}) are smaller for the comparison with $X_{thr\_gr}$. This is because both the multi-class VIM values and the discriminatory VIM values for the $X_{cl\_as\_4}$ covariates are larger than those for the $X_{thr\_gr}$ covariates.

For both the comparison of the $X_{cl\_as\_1}$ covariates with the $X_{two\_gr}$ covariates and with the $X_{thr\_gr}$ covariates, subtracting the corresponding discriminatory VIM values resulted in large AUC values for all MuF variants except \rcode{wosquared\_wgini}. As for $C=4$ and $C=6$, this is because, for these variants, the decrease between $X_{two\_gr}$ and $X_{cl\_as\_1}$ and between $X_{thr\_gr}$ and $X_{cl\_as\_1}$ was weaker for the multi-class VIM than for the discriminatory VIM (Figures~S10 to S13 in the supplementary material).

The exact AUC values shown in Figures~\ref{fig:AUC_C4} to \ref{fig:AUC_C10} are listed in Tables~S4 to S8 (supplementary material), with 95\% confidence intervals. These intervals are narrow and never wider than 0.05.

\myheading{Summary and conclusions}
To summarize, the multi-class VIMs typically ranked the class-associated covariates highest. This is unlike the conventional VIMs, which ranked the informative covariates without class-specific influence either similarly high or higher. This pattern held except in the setting with the smallest number of classes. Another exception was the setting with the smallest sample size ($n=100$), where the multi-class VIMs did not reliably rank the class-associated covariates highest.

The two versions \rcode{wsquared\_wgini} and \rcode{wsquared\_wogini} performed better than the other versions. This suggests that squaring the proportions $p_{c, \mathcal{I}^{as}_{m,c}}$ is important to reliably identify class-associated covariates. In hindsight, this result is not surprising. By definition, class-associated covariates have regions that substantially contain observations from specific classes. Squaring the proportions $p_{c, \mathcal{I}^{as}_{m,c}}$ rewards these regions more than when the proportions are not squared.

Comparing the multi-class VIM values with the discriminatory VIM values can help identify covariates that have relatively small multi-class VIM values but are still class-associated. Such covariates also have correspondingly small discriminatory VIM values. An example is $X_{cl\_as\_1}$ in Figure~\ref{fig:vimvalues}. Additionally, comparing the discriminatory VIM values can identify covariates that, while having relatively large multi-class VIM values, exhibit particularly large discriminatory VIM values, indicating they are only slightly class-associated. Examples in Figure~\ref{fig:vimvalues} are $X_{two\_gr}$ for $C=6$ and $C=10$ and $X_{thr\_gr}$ for $C=10$.

However, comparing with the discriminatory VIM values is not always informative. Class-associated covariates can also have high values with respect to the discriminatory VIM. For example, in Figure~\ref{fig:vimvalues} for $C=6$ and $C=10$, the class-associated covariates with the strongest influences have both the largest multi-class VIM values and the largest discriminatory VIM values among all covariates.

The results did not indicate that the split criterion associated with the variant \lq\lq Assign Class'' for the binary splits would lead to discriminatory VIM values more similar to the multi-class VIM values than the classic Gini split criterion. On the contrary, with the \lq\lq Assign Class'' split criterion, the discriminatory VIM values were systematically larger than the multi-class VIM values. Conversely, using the Gini split criterion for the binary splits resulted in discriminatory VIM values that were higher for some covariates and lower for others compared to the multi-class VIM values. This variability could provide clearer insights into which covariates are class-associated and which are influential but not class-associated. However, it is unclear to what extent the observation that the discriminatory VIM values are larger for some covariates and smaller for others than the multi-class VIM values is generalizable. Therefore, direct comparisons between the multi-class VIM values and the discriminatory VIM values should always be made with caution to avoid overinterpretation. In case of doubt, interpretation should be limited to the multi-class VIM values if the goal is to identify covariates with the strongest class-associated effects.

Due to the better performance of the variants \lq\lq Squared'' and \lq\lq Gini'' observed in this simulation study, we recommend the version \rcode{wsquared\_wgini} for identifying class-associated covariates.

\section{Real data analysis: Comparison of the predictive performance of multi forests with that of conventional random forests}
\label{sec:rdanalysis}

As outlined in the introduction, the primary goal of developing the MuF algorithm was to realize the multi-class VIM. It was not anticipated that MuFs would surpass conventional RFs in terms of predictive performance.
 
Nevertheless, it is reasonable to expect that users will employ MuFs not only to identify class-associated covariates but also for prediction purposes. Consequently, if MuFs demonstrate substantially inferior predictive performance compared to conventional RFs, caution should be advised when using them for prediction. Therefore, we investigated the predictive performance of MuFs, in addition to evaluating the performance of the multi-class VIM and the associated discriminatory VIM. Our analysis involved 121 real datasets with multi-class outcomes. In this study, we compared the predictive performance of the four versions of MuFs, as described in Section~\ref{sec:mufconstr}, with that of conventional RFs across four different performance metrics.

\subsection{Data}
\label{sec:data}

The datasets were primarily obtained from the Open Science online platform OpenML \citep{Vanschoren:2013}. We generally searched for datasets with multi-class outcomes, but in particular made sure that we included all suitable datasets from the curated dataset collection OpenML-CC18 \citep{Bischl:2019}. These were complemented by appropriate datasets from the Penn Machine Learning Benchmarks (PMLB) collection \citep{Romano:2021}.

Initially, datasets were filtered to include those with three to 20 outcome classes, a maximum of 15000 observations, between two and 500 covariates, and no missing values. The restrictions on the number of observations and covariates served to limit the computational burden. The datasets that passed these initial filters were then manually curated. Here we excluded redundant and many simulated datasets. However, we kept those simulated datasets whose descriptions indicated that they realistically represented the properties of real data. There were often groups of related, very similar datasets in the collection. In general, we selected only one dataset from each such group. Exceptions were groups of related datasets in which the datasets involved appeared adequately different. Furthermore, datasets with repeated measurements were excluded.

The selected datasets were pre-processed by removing categorical covariates with more than 50 categories, as well as overtly non-informative covariates such as subject IDs. In some cases, categorical covariates were cast as metric covariates in R following download from OpenML. We converted these covariates into  factors. This step was necessary for the MuF and RF implementations to correctly identify them as categorical covariates. Further details on the preprocessing are provided in the corresponding commented R script in the code publicly available on GitHub.

Ultimately, 121 datasets were included in the comparison study. Detailed information about these datasets, such as the number of observations, covariates, and outcome classes, is available in Tables~S9 to S11 (supplementary material).

Despite meticulous efforts to maintain high dataset quality, it cannot be guaranteed that all datasets meet stringent quality standards. However, given the large number of datasets, and our approach to interpreting results in aggregate rather than individually, the likelihood that minor quality issues have influenced our overall findings is minimal.

\subsection{Study design}
\label{sec:study_design}

Using the 121 datasets described above, we compared the predictive performance of the four versions of the MuF algorithm outlined in Section~\ref{sec:mufconstr} with that of conventional RFs. 

For the MuFs, as in the simulation study (Section~\ref{sec:simstudy}), subsampling with a proportion of $prop = 0.7$ was used instead of bootstrap for sampling observations for tree training. In each split, $npervar=5$ multi-way splits were again evaluated per sampled covariate and $mtry$ was again set to $\lfloor \sqrt{p} \rfloor$, with $p$ representing the number of covariates in the respective dataset. To limit the computational burden, the number of trees per forest, $ntree$, varied with the dataset size: 5000 trees for datasets up to 5000 observations (86\% of all datasets) and 1000 trees for datasets exceeding 5000 observations.

The configuration of the RFs was again adapted to that of the MuFs to avoid confounding effects due to different method configurations. The same numbers of trees were used as with MuFs and $mtry$ was set to $\lfloor \sqrt{p} \rfloor$. Subsampling with $prop = 0.7$ was again used instead of bootstrap. Additionally, as with MuFs, the categories of unordered categorical covariates were sorted using the PCA-based approach by \citet{Coppersmith:1999}.

In the presentation of the results, the different versions of the MuF algorithm are labeled as \rcode{wsquared\_wgini}, \rcode{wosquared\_wgini}, \rcode{wsquared\_wogini}, and \rcode{wosquared\_wogini}, consistent with the previous section. The conventional RFs are also labeled in typewriter font as \rcode{RF} for consistency.

We used the following performance metrics: AUNU, AUNP \citep{Fawcett:2001}, the Brier score \citep{Brier:1950}, and the accuracy. Both AUNU and AUNP are based on the AUC. For both measures, $C$ AUC values are summed up, where for $c \in \{1, \dots, C\}$ the $c$-th AUC value is calculated for the classification problem \lq\lq class $c$ yes vs.\ class $c$ no''. The difference between the two measures is in the weighting of the AUC values. For the AUNP, the AUC values are weighted with the corresponding class frequencies. In contrast, when calculating the AUNU, they are weighted with $1/C$, meaning that the simple average of the AUC values is formed. The AUNP therefore weights large classes more heavily, while the AUNU weights all classes equally, regardless of how many observations they are represented by. 

The Brier score measures the quality of probability predictions in relation to the true classes. Therefore, the Brier score is a measure of the calibration of models. However, it is also a measure of the discrimination in the sense that it measures how well observations of different classes can be correctly distinguished from each other by the evaluated model. The AUC, on the other hand, is only a measure of discrimination, but not of calibration. The accuracy should be interpreted with caution, as its value can depend heavily on the class distribution. For all forest types, the classes with the highest predicted probabilities were taken as the predicted classes when calculating the accuracy. For brevity, the Brier score and accuracy are hereafter referred to as \lq\lq Brier'' and \lq\lq ACC'', respectively.

The performance metrics were estimated for each dataset-method combination using 5-fold stratified cross-validation, repeated five times.

\subsection{Results}
\label{sec:results}

\begin{figure}[ht!]
\begin{center}
\includegraphics[width = \textwidth]{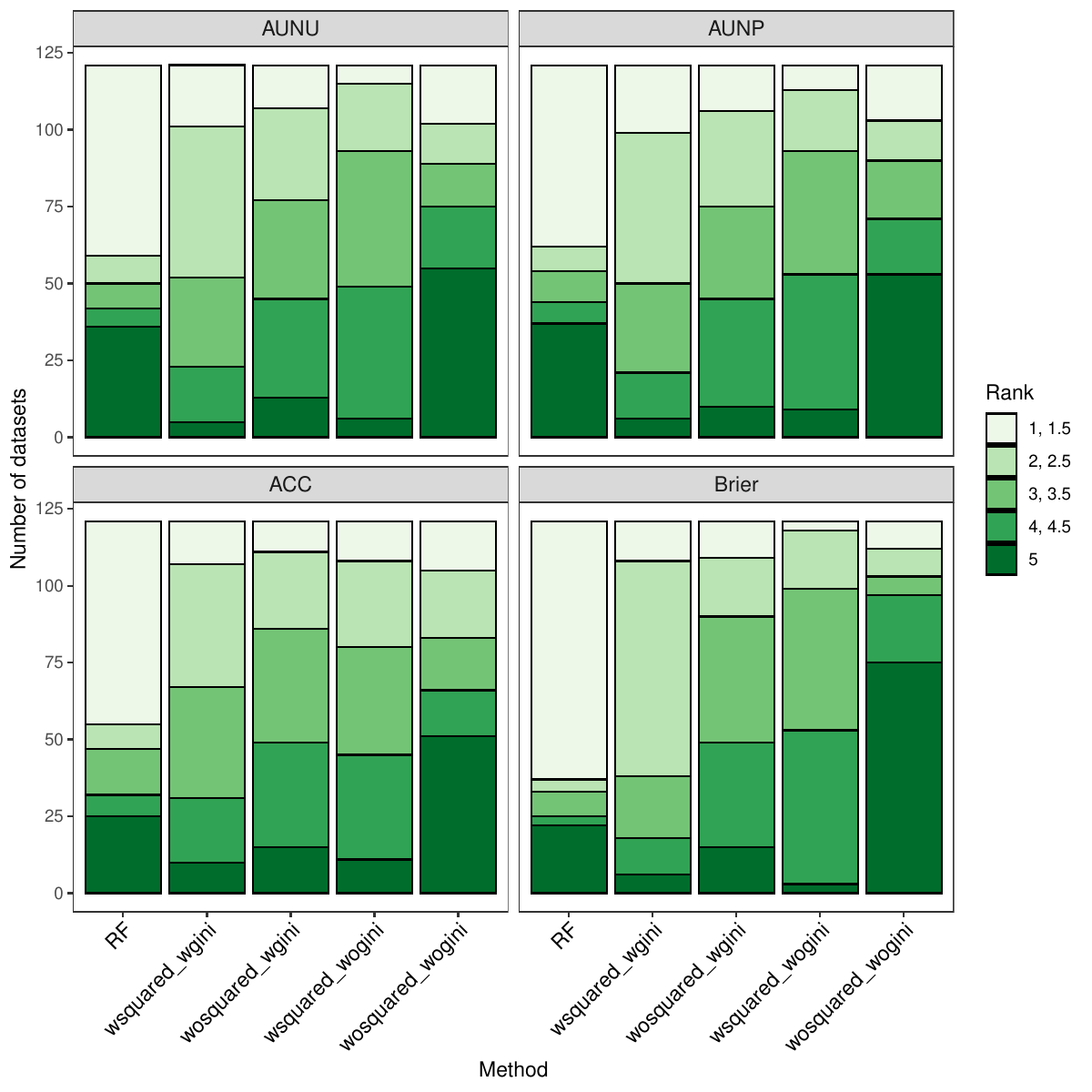}
\end{center}
\caption{Ranks of the MuF versions and \rcode{RF} with respect to the different performance metrics. Each stacked bar represents the number of datasets for which the respective method achieved the indicated ranks among all other methods.}
 \label{fig:benchmark_study_ranks}
\end{figure}

Table~\ref{tab:benchmark_summary} presents the medians of the cross-validated performance metrics obtained by the methods across all 121 datasets, along with the corresponding first and third quartiles. The differences in these median values are generally small. However, \rcode{RF} consistently shows the best performance, except in the case of AUNU. For AUNU, \rcode{wsquared\_wgini} had a slightly higher median value than \rcode{RF}.\footnote{However, Figure~\ref{fig:benchmark_study_ranks} discussed below does not support the suggestion that \rcode{wsquared\_wgini} outperforms \rcode{RF} concerning the AUNU.} For the other performance metrics, \rcode{wsquared\_wgini} ranked as the second-best, with the median of the ACC values for \rcode{wsquared\_wgini} matching that of \rcode{wsquared\_wogini}. Apart from this, the rankings among the medians achieved with the different MuF versions are not entirely consistent across the different performance metrics. The difference in performance between the different MuF versions and \rcode{RF} is more marked than between the MuF versions themselves. Although the differences among the methods are mainly small, the difference between \rcode{RF} and the MuF versions is notably greater for Brier. The rankings between the methods with regard to the first and third quartiles are largely consistent with those with regard to the medians.

\begin{table}[h]
\caption{Performance of the methods summarized over the 121 datasets. The values outside the parentheses show the medians of the cross-validated performance metrics, and the values inside the parentheses show the corresponding first and third quartiles (25\% and 75\% quantiles). Except in the case of Brier, larger values indicate better performance.}\label{tab:benchmark_summary}
\begin{tabular}{lllll}
  \hline
method & AUNU & AUNP & Brier & ACC \\ 
  \hline
RF & 0.9537 & 0.9571 & 0.3039 & 0.8159 \\ 
	 & [0.7860, 0.9972] & [0.7866, 0.9972] & [0.0960, 0.4927] & [0.6187, 0.9616] \\ 
wsquared\_wgini & 0.9575 & 0.9534 & 0.3353 & 0.8111 \\ 
& [0.7876, 0.9973] & [0.7836, 0.9970] & [0.1253, 0.5128] & [0.6192, 0.9589] \\ 
  wosquared\_wgini & 0.9570 & 0.9528 & 0.3365 & 0.8110 \\ 
 & [0.7852, 0.9973] & [0.7785, 0.9970] & [0.1258, 0.5126] & [0.6181, 0.9580] \\ 
  wsquared\_wogini & 0.9572 & 0.9531 & 0.3394 & 0.8111 \\ 
	 & [0.7832, 0.9970] & [0.7782, 0.9970] & [0.1261, 0.5136] & [0.6201, 0.9549] \\ 
  wosquared\_wogini & 0.9542 & 0.9523 & 0.3378 & 0.8103 \\ 
  & [0.7885, 0.9969] & [0.7770, 0.9969] & [0.1270, 0.5173] & [0.6131, 0.9544] \\ 	
   \hline
\end{tabular}
\end{table}

To obtain a more detailed insight into differences in performance between the methods, we examine Figure~\ref{fig:benchmark_study_ranks}. This figure visualizes the ranks the methods achieved for each dataset compared to the other methods, where lower ranks indicate better relative performance. It is evident across all performance metrics that \rcode{RF} achieved the best predictive performance most frequently. This is particularly evident for Brier, where \rcode{RF} delivered the best predictions for almost 70\% of the datasets. Additionally, this analysis confirms that \rcode{wsquared\_wgini} consistently provided the second-best results. Across all performance metrics, \rcode{wsquared\_wgini} delivered the best or second-best results more often than the other MuF versions. Notably, \rcode{RF} not only achieved the best results most frequently but also frequently the worst results. \rcode{RF} performed worst across all performance metrics more often than the MuF versions, with the exception of \rcode{wosquared\_wogini}. The latter version most frequently delivered the worst results for all performance metrics and generally ranked worst among the compared methods. The differences in the rankings between \rcode{wosquared\_wgini} and \rcode{wsquared\_wogini} are small; however, the former generally yielded marginally better results than the latter.

Given that \rcode{wsquared\_wgini} yielded the best predictive performance and \rcode{wosquared\_wogini} the worst, the variants \lq\lq Squared'' and \lq\lq Gini'' seem beneficial not only in identifying class-associated covariates (Section~\ref{sec:simstudy}) but also in prediction. Thus, \rcode{wsquared\_wgini} is the sole MuF version recommended and is the only version available in the R package \rcode{diversityForest}, which implements MuFs.

We tested for differences between the performance metric values obtained with \rcode{RF} and the MuF versions. However, we did not test for differences among the MuF versions. This approach reduced the number of comparisons and, consequently, the number of p-values requiring adjustment for multiple testing. As a result, the power of the tests was increased over what would have been achieved with more comparisons. Furthermore, it is of minimal relevance to determine whether the predictive performance of the different MuF versions is statistically significantly different. As noted above, these four versions are alternatives, and only one was recommended for application. The recommendation was based solely on the observed performance. It was not relevant here whether the observed performance of the selected version was statistically significantly different from that of the other versions.

We conducted paired Wilcoxon tests, treating the cross-validated performance metrics from the different datasets as independent observations. The four p-values obtained for each performance metric were then adjusted for multiple testing using the Holm-Bonferroni procedure. Table~\ref{tab:wilcox_test} presents the test results. Except for the comparison between \rcode{wsquared\_wgini} and \rcode{RF} for AUNU and AUNP, all differences tested are statistically significant. The effect size values, $r$, for AUNU, AUNP, and ACC ranged from small to medium, while those for Brier indicated larger performance differences between the MuF versions and \rcode{RF}.

\begin{table}[h]
\caption{Results of Wilcoxon tests between the values of the cross-validated metrics obtained with the different MuF versions and \rcode{RF}. The values outside and inside of the round brackets represent the effect sizes $r$ and the Holm-Bonferroni adjusted p-values of the tests, respectively.}\label{tab:wilcox_test}
\begin{tabular}{lllll}
  \hline
method & AUNU & AUNP & ACC & Brier \\ 
  \hline
wsquared\_wgini & 0.14 (0.134) & 0.10 (0.287) & 0.27 (0.003) & 0.55 ($<$ 0.001) \\ 
  wosquared\_wgini & 0.27 (0.006) & 0.25 (0.015) & 0.35 ($<$ 0.001) & 0.58 ($<$ 0.001) \\ 
  wsquared\_wogini & 0.29 (0.004) & 0.26 (0.015) & 0.31 (0.001) & 0.60 ($<$ 0.001) \\ 
  wosquared\_wogini & 0.38 ($<$ 0.001) & 0.33 (0.001) & 0.40 ($<$ 0.001) & 0.63 ($<$ 0.001) \\ 
   \hline
\end{tabular}
\end{table}

We acknowledge that the p-values listed in Table~\ref{tab:wilcox_test} might be slightly underestimated. This is due to the fact that, as outlined in Section~\ref{sec:data}, there are occasionally groups among the datasets used that are related for different reasons. The datasets within these groups are therefore not completely independent. However, as described in Section~\ref{sec:data}, we selected only one dataset from groups containing very similar datasets, which should minimize dependencies. Moreover, the p-values are in most cases substantially lower than the significance level $\alpha = 0.05$, making the presence of false positives in these results unlikely.

\begin{figure}[ht!]
\begin{center}
\includegraphics[width = \textwidth]{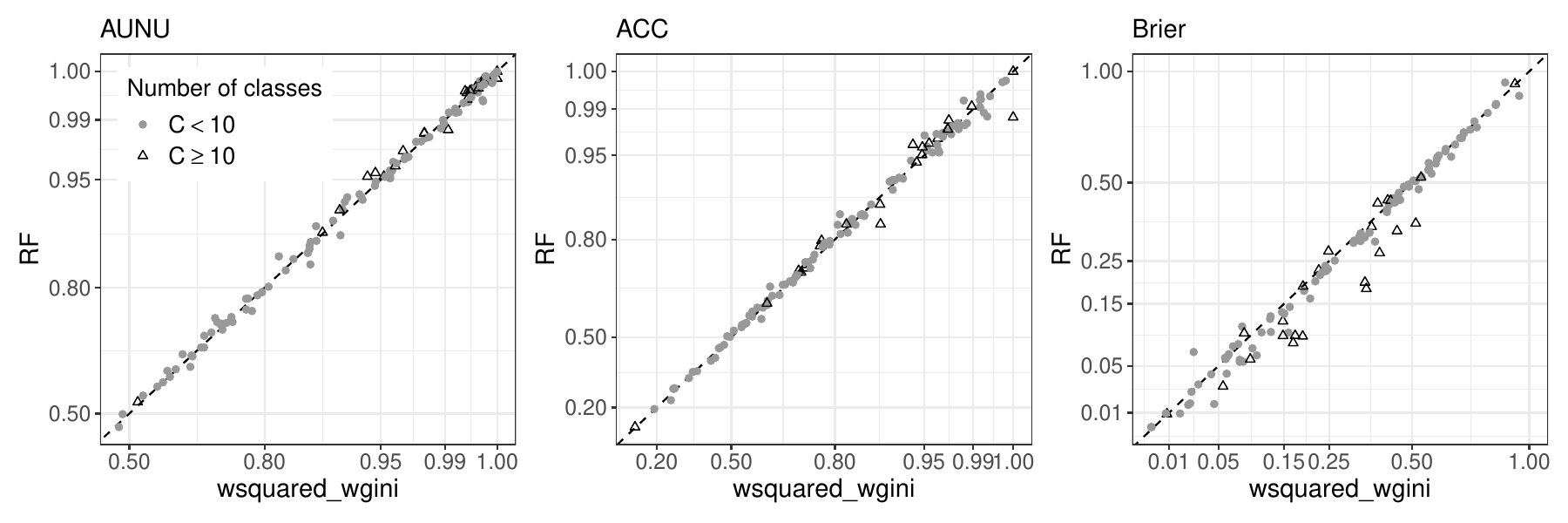}
\end{center}
\caption{Dataset-specific cross-validated performance metric values: \rcode{wsquared\_wgini} versus \rcode{RF}. The white diamonds mark the datasets that have ten or more outcome classes. The dashed lines represent the diagonal. Axes were transformed for visual clarity: A negative complementary square root transformation was applied to AUNU and ACC, and a square root transformation was used for Brier. AUNP is not included here due to its results being very similar to AUNU and for space considerations. Figure~S16 (supplementary material) shows a version of this figure that includes AUNP.}
 \label{fig:metrics_wsquared_wgini_vs_RF}
\end{figure}

In Figure~\ref{fig:metrics_wsquared_wgini_vs_RF}, we plot the performance metric values of \rcode{wsquared\_wgini} and \rcode{RF} obtained for the individual datasets against each other. The correlation between these values is consistently strong. For AUNU and ACC, the observed differences between the two methods are negligible. However, consistent with the Wilcoxon test results, the differences for Brier are more pronounced, especially in datasets containing many classes ($C \geq 10$). Figure~S15 (supplementary material) is a version of Figure~\ref{fig:metrics_wsquared_wgini_vs_RF} in which we present the results separately according to whether the datasets contain categorical covariates or not. This figure indicates that the larger Brier differences predominantly occured in datasets lacking categorical covariates. Potential reasons for this observation will be explored in the subsequent discussion section.

\section{Discussion}
\label{sec:discussion}

\myheading{Overview of the proposed methodology and the results of the comparative studies}
In this paper, we have presented and evaluated a variant of RFs, named MuFs, tailored for datasets with multi-class outcomes in conjunction with two VIMs: the multi-class VIM and the discriminatory VIM. Unlike conventional RFs that use binary splitting only, MuFs in addition employ multi-way splits, dividing the nodes into more than two child nodes. The split criterion used here is aimed at finding multi-way splits where each child node substantially comprises observations from a different class.

The objective of the multi-class VIM is to identify what we termed class-associated covariates. These covariates are specifically linked to one or more classes and are characterized by having value regions where these classes are prominently represented. Conventional VIMs tend to also highly rank covariates that merely differentiate between groups of classes rather than being associated with specific classes. These covariates should be assigned lower importance scores by the multi-class VIM. On the other hand, like conventional VIMs, the discriminatory VIM aims to highly rank influential covariates of any type, irrespective of their association with specific classes.

In our simulations, the multi-class VIM ranked class-associated covariates higher than other influential covariates in the great majority of cases, except in the case of the smallest sample size considered. In contrast, the conventional VIMs generally failed to differentiate class-associated covariates from other influential covariates. An exception occurred with class-associated covariates with very strong influences; these were ranked highest by all VIMs evaluated. The discriminatory VIM displayed behavior very similar to that of the conventional VIMs.

The extensive real data analysis revealed that in many cases MuFs have a slightly inferior predictive performance compared to conventional RFs. The differences were minor in terms of accuracy and the performance metrics AUNU and AUNP, which assess discrimination ability. Slightly larger differences were observed for the Brier score, which additionally evaluates model calibration. Consequently, the probability predictions made by MuFs appear to be slightly less well calibrated than those of conventional RFs, without this considerably impairing their classification performance.

\myheading{Importance of interpretability and problem-specific predictive performance}
The primary focus of the MuF algorithm is, however, not prediction but rather interpretation, facilitated by the multi-class VIM and the discriminatory VIM. Consequently, the fact that MuFs often do not surpass conventional RFs in predictive performance is not a particular limitation. Many classification and regression methods remain widely used in practice despite evidence that other methods often yield superior prediction results. An important reason for this is that, like MuFs, these methods often provide benefits in terms of interpretability. For instance, classical generalized linear models offer clear, easily communicable covariate effects, which is a substantial advantage over many contemporary machine learning models.

Another reason for the continued use of methods that achieve better prediction results less frequently than other methods is the \lq\lq No Free Lunch'' theorem. According to this theorem, no classification or regression method can universally perform best across all possible prediction problems \citep{Wolpert:1997}. This highlights the importance of selecting methods based on context and their specific performance in the prediction problem being addressed.

In scenarios where targeting the highest possible predictive performance is critical, we recommend considering alternative methods in addition to MuFs, such as conventional RFs.

\myheading{Suboptimal predictive performance: Possible reasons and problems for large numbers of classes}
One potential cause of MuFs' suboptimal predictive performance is the fact already mentioned in the introduction that multi-way splits partition the data recursively very quickly. This could result in fewer covariates being used within each tree compared to conventional RFs and the interactions between the covariates being less exploited. Although incorporating binary splits in the MuF algorithm was intended to mitigate these issues, it likely did not fully resolve them.

These challenges are likely especially pronounced in scenarios with large numbers of outcome classes. In such cases, each node is divided into many child nodes, leading to particularly rapid data partitioning. This was also reflected in the real data analysis (Section~\ref{sec:rdanalysis}), where datasets with the strongest declines in MuFs' predictive performance relative to RFs typically had many outcome classes ($C \geq 10$). Consequently, for prediction problems involving 10 or more outcome classes, we recommend using the MuF algorithm solely to compute the multi-class and discriminatory VIM values. For prediction, conventional RFs or other appropriate methods should be used in such scenarios. Note also that the MuF algorithm was not designed for scenarios involving very large numbers of outcome classes. This is also the reason why only datasets with up to 20 outcome classes were included in the real data analysis.

Furthermore, in the MuF algorithm, the split criterion used for multi-way splits that involve a number of unique covariate values at least as great as the number of outcome classes exclusively rewards class-associated covariates. This likely results in an under-selection of other influential covariates for such splits. In contrast, conventional RFs generally select influential covariates, irrespective of class association. This likely results in conventional RFs more effectively leveraging the available predictive information within the covariates.

In the real data analysis, we observed that the predictive performance of MuFs was worse relative to that of conventional VIMs in datasets lacking categorical covariates. This may be attributed to the general potential reasons for the suboptimal predictive performance of MuFs discussed above. First, it is to be expected that the recursive partitioning of the data is particularly fast when only continuous covariates are present. This is because multi-way splits based on continuous covariates generally result in more child nodes than those based on categorical covariates. Second, the advantage of conventional RFs that they take into account influential covariates, also if they are not class-associated, can be expected to diminish as the proportion of categorical covariates increases. This is because, for categorical covariates, the MuF algorithm generally does not employ the split criterion that rewards only class-associated covariates. Consequently, the limitation of MuFs in exploiting the predictive information in the covariates less effectively, becomes less pronounced when more categorical covariates are present.

\myheading{The importance of class-specific intervals for interpretability}
In the multi-way splits of MuFs, the covariate values are grouped into adjacent intervals corresponding to different classes. Therefore, only those class-associated covariates where one or more intervals contain numerous observations of different classes are adequately considered. Covariates with multiple intervals dominantly representing the same classes are likely less prioritized. 

Given that the multi-class VIM is based on the multi-way splits, it is thus likely that the last-mentioned types of covariates, hereafter referred to as \lq\lq type $\mathrm{II}$ covariates'', generally receive lower multi-class VIM values than the first-mentioned types of covariates, hereafter referred to as \lq\lq type $\mathrm{I}$ covariates''. This distinction is beneficial for interpretability because type $\mathrm{I}$ covariates have more easily interpretable influences. For instance, the influence of a type $\mathrm{I}$ covariate could be the following: For small values of the covariate, class $A$ may be likely, for slightly larger values, however, class $B$ may be more likely and for large values, class $C$ might be most likely. In contrast, a possible influence of a type $\mathrm{II}$ covariate could be the following: For small or very large values of the covariate, class $A$ may likely, while for moderate or large values, classes $B$ or $C$ might be more likely, respectively.

\myheading{Reason for not optimizing the tuning parameters}
Similar to conventional RFs, the MuF algorithm incorporates several tuning parameters. To limit the computational effort, these were not optimized in our comparative studies. This approach is supported by the findings of \citet{Probst:2019} based on many datasets, which suggest that the performance of RFs is only slightly affected by the tuning parameter values. Since MuFs are special cases of diversity forests, which in turn are a subclass of RFs, it is reasonable to assume similar findings apply to MuFs. In addition, the influence of the parameter $nsplits$ on the performance of diversity forests was investigated in \citet{Hornung:2022b} in an extensive empirical analysis. Here, similar to with conventional RFs, the influence of this parameter was very small for the majority of the datasets used. Note that the parameter $nsplits$ is referred to as $mtry$ in the MuF algorithm, deliberately mirroring the terminology used in conventional RFs to highlight the similarities of these parameters.

In the analyses reported in this paper, for most tuning parameters we chose default values based on empirical evidence-based recommendations from existing literature on conventional RFs (subsampling, $prop$, $mtry$). As outlined in Section~\ref{sec:comp_methods}, the default values for the parameters $ntree$ and $npervar$ were determined using some of the datasets from the real data analysis. However, it can be safely assumed that this did not lead to an optimistic bias in the results. The selection of the default values was not based on the predictive performance nor on the VIM values obtained. It was based solely on the observed stability of the VIM values obtained.

\myheading{Identification of covariates associated with all classes: Conceptual difficulties and possible realization in the MuF algorithm}
The aim of the multi-class VIM is to identify covariates specifically associated with one or more classes. By contrast, the approaches of \citet{Zini:2015} and \citet{Song:2017}, as discussed in Section~\ref{sec:relwork}, focus on identifying covariates associated with all classes. Consequently, the distributions of these covariates exhibit differences between all classes.

However, the intention to exclusively search for such covariates might not always be reasonable. One reason is that these covariates often represent only a small proportion of all informative covariates. Additionally, for covariates that do exhibit differences across all classes, these differences are unlikely to be large if they are only detectable through specialized procedures. This is because covariates showing notably strong differences across all classes typically exert such a strong influence that they are already highly ranked by conventional VIMs. Conversely, if covariates show only minor differences across all classes, their influence is likely not of substantive interest.

For scenarios where it is still relevant to identify exclusively covariates with differences between all classes, a potential modification of the MuF algorithm could be developed. In this modification, both the multi-way splitting criterion and the calculation of the multi-class VIM might be based on the minimum of the squared proportions $p_{c,\mathcal{I}^{as}_{m,c}}^2$, rather than on their sum. By maximizing the minimum proportions, specific types of multi-way splits would be favored. Such splits would ensure that none of the resulting intervals contain merely a few observations from the classes assigned to the respective child nodes. However, this potential modification of the MuF algorithm has not been explored, and thus no assertions can be made about its efficacy.

\myheading{Graphical exploration of covariate influences to supplement the multi-class VIM analysis}
Whether the goal is to identify covariates associated with all classes or, more generally, covariates associated with one or more specific classes, multi-class VIM values do not provide information about the forms of the influences of these covariates. To gain insights into how these covariates impact the outcome, graphical representation techniques can be used. The \rcode{diversityForest} R package includes functions that estimate and visualize the dependency structures of the multi-class outcome on the covariates. Each covariate of interest is represented in a separate plot. The package offers two types of visualizations: kernel density estimate-based plots and boxplot-based plots.

Kernel density estimate-based plots display the within-class distributions of the covariate in a single graph, where the density estimates are scaled according to the class proportions to account for unequal class sizes. These plots facilitate learning which classes are most dominant in different covariate value regions. On the other hand, boxplot-based plots present the estimated within-class distributions side by side, using boxplots. This type of visualization makes it easier to determine where in the covariate values most of the observations from each class are concentrated, which can be less apparent in kernel density estimate-based plots, especially when the number of classes is large. However, unlike kernel density estimates, the boxplot-based plots often do not allow to judge which classes dominate specific regions.

We strongly encourage interested readers to consult the illustrative analyses in the example sections of the plotting functions within the \rcode{diversityForest} R package. These examples demonstrate how these visualizations can be used in a multi-forest analysis.

\section{Conclusions}
\label{sec:conclusions}

MuFs represent a variant of RFs tailored for multi-class outcomes. The most important aspect of the MuF algorithm, however, is not the predictive models it can create, but the multi-class VIM. This tool enables the specific identification of class-associated covariates, which are distinguished by their specific association with one or more classes. Conventional VIMs do not allow this, as they target influential covariates indiscriminately without focusing on class association. Similarly, the discriminatory VIM, also part of the MuF algorithm, allows for the identification of influential covariates of any type.

Compared to conventional RFs, MuFs frequently demonstrate slightly lower predictive performance, particularly in terms of calibration and for datasets with many outcome classes. Therefore, for prediction, other methods besides MuFs, such as conventional RFs, should be considered in the following situations: 1) when it is critical to attain the maximum possible predictive performance, 2) when dealing with datasets with many outcome classes.

However, the primary purpose and distinctive feature of MuFs lie in their ability to specifically identify class-associated covariates using the multi-class VIM. Given this specialized functionality, the often slightly reduced predictive performance compared to conventional RFs does not represent a particular limitation.

\section*{Acknowledgements}
This work was supported by the German Science Foundation [DFG-Einzelförderung  HO 6422/1-3 to Roman Hornung]. It is also related to the COAT project funded by the German Science Foundation [DFG, Projektnummer (grant number): 447467169 to Alexander Hapfelmeier].

\section*{{S}upplementary material and R code}
The supplementary material can be found at the following link:\\
\url{https://www.ibe.med.uni-muenchen.de/organisation/mitarbeiter/070_drittmittel/hornung/multiforests_suppfiles/suppmat_multiforests.pdf}. The R code used to produce the results shown in the main paper and in the supplementary material is available on GitHub (\url{https://github.com/RomanHornung/MultiForests_code_and_data}).

\bibliographystyle{abbrvnat}
\bibliography{bibliography}

\end{document}